\documentclass[lettersize,journal]{IEEEtran}
\usepackage{amsmath,amsfonts}
\usepackage{textcomp}
\usepackage{stfloats}
\usepackage{url}
\usepackage{verbatim}
\usepackage{graphicx}
\usepackage{cite}
\usepackage{amsmath,amsfonts}
\usepackage{algorithmic}
\usepackage{algorithm}
\usepackage{array}
\usepackage[caption=false,font=footnotesize,labelfont=rm,textfont=rm]{subfig}
\usepackage{bbding}
\usepackage{booktabs}
\usepackage{multirow}
\usepackage{makecell}
\usepackage{amsmath,amsthm,amssymb,amsfonts}
\usepackage{threeparttable}
\usepackage{xcolor} 
\usepackage{tcolorbox}
\usepackage[colorlinks=true, linkcolor=blue, citecolor=blue, urlcolor=blue]{hyperref}

\begin{document}

\title{Resilient Concurrent Causal Discovery for Topological Event Sequences}

\author{Jiyu Tian,~Junhao Dong,~Mingchu Li,~Lingling Fang,~Liming Chen,~Andreas Holzinger,~Zheng Yan,~Ong Yew Soon


\thanks{Jiyu Tian and Mingchu Li are with the School of Software Technology, Dalian University of Technology, Dalian, China. Mingchu Li is also affiliated with the School of Computer and Information Engineering, Jiangxi Normal University, Nanchang, Jiangxi, China. Email: tianjiyu@mail.dlut.edu.cn, mingchul@dlut.edu.cn}

\thanks{Liming Chen is with the School of Computer Science and Technology, Dalian University of Technology, Dalian, China. Email: limingchen0922@dlut.edu.cn}

\thanks{Lingling Fang is with the School of Computer Science and Artificial Intelligence, Liaoning Normal University, Dalian, China. Email: fanglingling@lnnu.edu.cn}

\thanks{Andreas Holzinger is with the Human-Centered AI Lab, FTEC, Department of Ecosystem Management, Climate and Biodiversity, University of Natural Resources and Life Sciences (BOKU), Vienna, Austria. Email: andreas.holzinger@human-centered.ai}

\thanks{Zheng Yan is with the State Key Lab of ISN, School of Cyber Engineering, Hangzhou Institute of Technology, Xidian University, Xi’an, Shaanxi, China. Email: zyan@xidian.edu.cn}

\thanks{Junhao Dong and Yew Soon Ong are with the College of Computing \& Data Science, Nanyang Technological University, Singapore, and Yew Soon Ong also with CFAR, IHPC, A*STAR, Singapore. Email: junhao003@ntu.edu.sg, asysong@ntu.edu.sg}
}

\markboth{ }
{How to Use the IEEEtran \LaTeX \ Templates}

\maketitle

\begin{abstract}
Causal discovery on topological event sequences is crucial for ensuring the reliability of networks. However, existing methods struggle to capture the complex causal relationships arising from concurrent events and lack robustness to incomplete event sequences. To address these issues, we propose a resilient concurrent causal discovery method, termed RCCD, enabling robust learning of causal graphs from topological event sequences. Specifically, we first introduce an influence-aware hyperedge causal attention mechanism, which incorporates event duration into the embedding representation, aggregates concurrent event features via hyperedge causal convolution, and injects network prior knowledge to capture the complex many-to-one causal interactions. Furthermore, we design a masked-based alternating causal optimization framework, which forces the model to recover masked event types based on context through self-supervised mask reconstruction, thereby enhancing the resilience of the predictor to missing data. To validate the effectiveness of our method, we conduct extensive experiments on both simulated and real-world telecommunication network datasets. Experimental results demonstrate that the proposed method significantly outperforms existing state-of-the-art methods in both accuracy and robustness, making it more suitable for real-world telecommunication network environments. 

\end{abstract}


\section{Introduction}
\label{sec:1}

\IEEEPARstart{T}{elecommunication} networks support numerous communication services and serve as critical infrastructure for modern society \cite{surveytelcom,surveytelcom2,tnt}. During network operations, anomalies such as device failures inevitably trigger alarm events, and log systems record these events in chronological order, forming discrete event sequences \cite{logparsing, logparsingsurveys, ssdalog}. Given the massive scale of alarm events, it is unreliable for operation and maintenance (O\&M) personnel to directly troubleshoot from the original sequences \cite{hfaw,loghub}. Event sequence causal discovery methods can identify causal relationships among event types from large-scale alarm sequences, helping O\&M personnel promptly locate root causes, block chain reactions, and trace anomaly propagation paths, thereby ensuring the stable operation of networks \cite{cdsurveys1, cdsurvery2, cdsurveys3}.

\begin{figure}[h]
  \centering
  \includegraphics[width=0.98\linewidth]{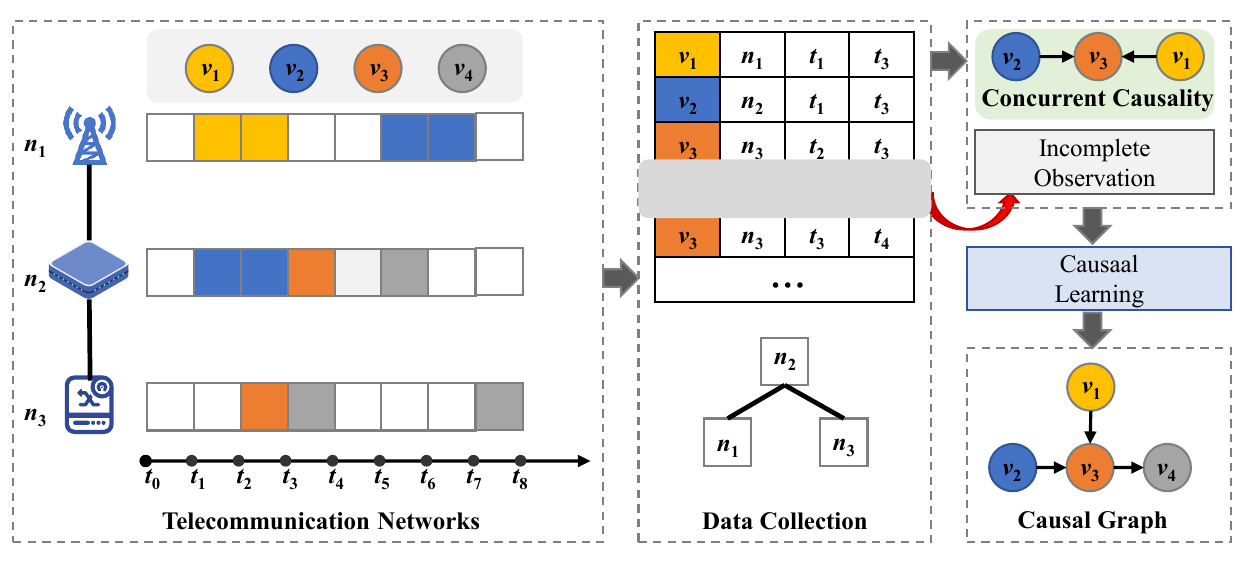}
  \caption{An illustrative example of topological event sequence causal discovery. Causal discovery for telecom network event sequences can identify causal relationships from discrete alarms generated by devices. Each alarm event includes an event type, a device type, start timestamps, and end timestamps. Network topology describes the physical connections among devices. Concurrent events refer to multiple alarms occurring simultaneously within the same time window, which may collaboratively produce complex causality. Incomplete observations mean that some alarms are not recorded due to network instability. Causal graph is a directed acyclic graph where nodes represent event types and directed edges indicate that one type directly triggers another type.}
  \label{figure1}
\end{figure}

Unlike general discrete event sequences, telecommunication network event sequences exhibit distinctive characteristics. Specifically, as shown in Figure \ref{figure1}, anomaly propagation is constrained by the network topology. An alarm on one device may trigger related alarms on neighboring devices, so event sequences generated by different devices are not independent and identically distributed (i.i.d). Second, in large networks, multiple alarms may occur simultaneously, and their durations affect the intensity of fault propagation, resulting in concurrent and complex causal relationships. Furthermore, networks typically consist of heterogeneous devices from multiple vendors, and the data collection process often suffers from event missing due to network congestion or parsing errors, leading to incomplete observed sequences. Therefore, causal learning approaches for telecommunication networks need to mine complex interactions among events from incomplete sequences under topological constraints.

Fortunately, for the non-i.i.d. sequence of telecommunication networks, existing methods have introduced topology-aware statistical point processes \cite{s2gcsl, tccd, statehps} and neural point processes for causal discovery. For example, statistical point process methods such as S$^2$GCSL \cite{s2gcsl}, and TCCD \cite{tccd} incorporate network topology as a prior constraint into Hawkes process-based causal modeling. Neural point process methods, such as THPs \cite{thps}, TNPAR \cite{tnpar}, and CausalNet \cite{causalnet}, utilize neural networks to capture temporal dependencies, optimizing the causal matrix as trainable weights alongside the neural network. However, these methods either fail to effectively handle the complex causal relationships of concurrent events or lack resilience to incomplete observations. We formalize the research challenges addressed in this paper as follows:

\begin{itemize}
\item \textbf{Complexity: }Existing methods are insufficient in capturing the complex causal relationships resulting from concurrent events. For example, the TNPAR \cite{tnpar} method performs temporal causal modeling based on a simple multi-layer perceptron, making it difficult to capture the complex interactions arising from concurrent and collaborative events. CausalNet \cite{causalnet} method computes similarities between events via standard self-attention, which overlooks the influence of event duration on anomaly propagation paths and cannot express the joint causal effect when multiple events are triggered simultaneously. Therefore, telecommunications networks urgently require a causal discovery method capable of addressing concurrent alarms.

\item \textbf{Robustness: }Existing methods lack robustness in coping with missing sequences during data collection. For instance, both statistical point process methods \cite{s2gcsl, tccd, statehps} and neural point process methods \cite{thps, tnpar, causalnet} almost uniformly assume that event sequences are complete, entirely ignoring scenarios with missing data. Although StateHPs \cite{statehps} method handles non-stationary sequences by first segmenting and then modeling each segment independently, its multi-step training process still relies on complete event sequences, and data missing can impair the performance of causal optimization. Consequently, the reliability of causal discovery methods needs further improvement to adapt to telecommunication networks with incomplete observations.
\end{itemize}

To address the challenges above, we propose RCCD, a robust causal discovery method for complex telecommunication network alarm event sequences, designed to efficiently capture concurrent causality under topological constraints while resisting the impact of incomplete observations. Specifically, to tackle the complexity challenge, we design an influence-aware hyperedge causal attention \textbf{(IHCA)} mechanism that encodes event durations, employs hyperedge causal convolution, and incorporates topological decay coefficients into attention weights to simultaneously handle temporally adjacent concurrent events. 
Furthermore, to address the robustness issue, we design a mask-based alternation causal optimization \textbf{(MACO)} framework based on the Expectation-Maximization (\textbf{EM}) algorithm. In the E-step, we randomly mask a portion of historical events and perform self-supervised prediction as a mask reconstruction loss to ensure the robustness of the predictor to missing data. In the M-step, we fix a predictor and optimize the causal graph parameters, leveraging a robust predictor for causal learning. Compared with existing methods \cite{pc, grandag, corl, thps, tnpar, s2gcsl, causalnet, tccd}, our approach achieves state-of-the-art (SOTA) performance in both concurrent event causal modeling and robustness to incomplete observations, providing a novel solution for causal learning from large-scale discrete event sequences. In summary, the main contributions of this paper are as follows:

\begin{itemize}
\item To the best of our knowledge, this is the first endeavor on event sequence causal learning designed for incomplete observations, enabling resilient causal discovery from missing event sequences.

\item We propose an influence-aware hyperedge causal attention mechanism that integrates event duration encoding, hyperedge causal attention score computation, and prior knowledge constraints to address complex causal relationships.

\item We design a mask-based alternation causal optimization framework that enhances the robustness of the predictor to incomplete observations through self-supervised reconstruction training.

\item We conduct extensive experiments on both synthetic datasets and real-world data, validating the effectiveness of the proposed method in terms of causal learning performance and robustness.
\end{itemize}

The remainder of this paper is organized as follows. Section~\ref{sec:2} reviews related work on causal discovery from event sequences. Section~\ref{sec:3} presents the problem definition. Section~\ref{sec:4} introduces the overall architecture and elaborates on two novel components, respectively. Section~\ref{sec:6} presents comprehensive empirical studies. Section~\ref{sec:7} discusses the effectiveness and limitations of RCCD. Section~\ref{sec:8} concludes the paper.

\section{Related Work}
\label{sec:2}
Existing studies on event sequence causal discovery have made progress, but they ignore missing events and concurrent causal relationships. Temporal causal attention mechanisms lack network topology priors and have limited adaptability to non-i.i.d. discrete sequences. Therefore, we survey related works from two aspects: event sequence causal learning and temporal causal attention mechanisms.

\subsection{Causal Learning on Event Sequences}
\label{sec:2.1}

Event sequence causal discovery utilizes discrete alarm events recorded in chronological order to infer causal relationships among event types \cite{mdl, causalformer, llmgc, semsin, tecdi}. Depending on whether network topology is incorporated into the causal learning process, existing methods can be classified into two categories: topology-agnostic methods \cite{busca,cause,l0hawkes, cutsplus,cuts2, mml,lcdbk,cagke} and topology-aware methods\cite{thps,tnpar,s2gcsl,tccd,anhp}.

Topology-agnostic causal discovery methods typically assume that event sequences are independent and identically distributed, ignoring the network adjacency relationships among devices. As a statistical causal discovery method, the PC algorithm \cite{pc} constructs a causal graph through conditional independence tests and searches for an optimal directed acyclic causal graph based on the Bayesian Information Criterion and the Minimum Description Length score. However, the PC algorithm requires numerous conditional independence tests, limiting its practicality. To address this issue, PCMCI+ \cite{pcmci} further improves the selection of conditioning sets, reducing the number of tests from exponential to polynomial, thereby enhancing the efficiency and reliability of causal discovery in highly autocorrelated time series while providing link confidence estimates. GraNDAG \cite{grandag} transforms discrete directed acyclic graph (DAG) search into a continuous optimization problem, utilizing a feedforward neural network to capture complex nonlinear relationships among variables, and combines this with a gradient-based smooth acyclic penalty mechanism for end-to-end learning. CORL \cite{corl} formulates causal graph search as a Markov decision process, uses an encoder-decoder to generate a topological ordering of variables, takes a score function as the reward signal, and optimizes the continuous acyclicity constraint via reinforcement learning. SHPs \cite{shps} method addresses the causal identifiability problem in discrete Hawkes processes by modeling the instantaneous effects of events. Building on SHPs, CASCADE \cite{cascade}, based on the algorithmic Markov condition, models causal discovery as an optimization problem of minimizing Kolmogorov complexity, instantiates the score function via the minimum description length principle, and incrementally adds causal edges in topological order, enabling it to handle both instantaneous and delayed effects simultaneously.

Topology-aware causal discovery methods embed the physical adjacency relationships between devices as prior constraints into the model, thereby restricting causal inference between event types to the network topology. Simultaneously, these methods can be further divided into statistical point process-based methods and neural point process-based methods. Among statistical point process methods, S$^2$GCSL \cite{s2gcsl} employs a linear kernel Hawkes process and stochastic gradient descent to efficiently optimize the likelihood function, using L$_1$ regularization to enforce sparsity. Following S$^2$GCSL, TCCD \cite{tccd} combines Monte Carlo tree search with continuous gradient optimization, performing global tree exploration in the causal graph space and local parameter tuning via continuous optimization, thereby effectively overcoming the tendency of pure gradient methods to get trapped in local optima. StateHPs \cite{statehps} handles non-stationary event sequences by first segmenting the sequence and then modeling each segment independently. However, its multi-step training process still relies on the completeness of events within each segment, and missing data directly disrupts the causal structure optimization within the segments. 
Among neural point process methods, THPs \cite{thps} is the first to incorporate network topology into event sequence causal discovery. By combining topological graph convolution with temporal convolution to enhance the Hawkes process, and employing the Expectation-Maximization algorithm with sparse regularization to search for causal structures, THPs enforces the propagation of event influences along actual physical links.
TNPAR \cite{tnpar} builds on THPs and proposes a topological neural Poisson autoregressive model, leveraging the powerful nonlinear representation capability of neural networks to capture complex dynamic evolution relationships among discrete events. CausalNET \cite{causalnet} introduces a causal weight matrix into the Transformer, injects physical topology constraints into causal attention computation, optimizes the causal graph using Gumbel-Softmax, and alternately optimizes the prediction module and the causal graph structure during training. 
Unlike these causal discovery methods, coevolutionary multitasking optimizes multiple tasks concurrently via cooperative coevolution \cite{ong1}, and co-evolutionary multi-task learning employs predictive recurrence for multi-step prediction \cite{ong2}.

Although existing methods have made significant progress in causal discovery from event sequences, they still lack exploration of missing events and have not fully considered many-to-one causality, that are prevalent in telecommunication networks.

\subsection{Temporal Causal Attention Mechanism}
\label{sec:2.2}

In the task of causal discovery from event sequences, existing methods utilize the powerful temporal dependency modeling capabilities of Transformers \cite{transformer, difftransformer,mltransformer} to learn causal graphs from observational sequences \cite{causalnet, causalformer}. As the core component of the Transformer, the self-attention mechanism mimics human selective attention by dynamically assigning weights to elements at different positions when processing input data, thereby focusing on important information \cite{cadn, hadamard, qksan}. Based on the temporality and causality of event sequences, we review existing attention works from two dimensions: temporal attention \cite{transformertssurvey, psttransformer, dlinear,depmstat,mo3tr,stman} and causal attention \cite{cdt4rec,nopes, ergo, catformer}.

Regarding the temporal attention mechanism, existing works have conducted extensive explorations based on standard attention. To simultaneously capture long-term and short-term dependencies, THP \cite{thp} combines the self-attention mechanism with the Hawkes process to specifically model discrete event sequences. Informer \cite{informer} proposes the ProbSparse self-attention mechanism, which evaluates the sparsity of query vectors based on Kullback-Leibler divergence and retains only a few dominant queries for dot-product computation to reduce computational complexity. Autoformer \cite{autoformer} proposes an autocorrelation attention mechanism that calculates autocorrelation coefficients using temporal periodicity to detect subsequence similarity, enabling the discovery of temporal relationships and information aggregation at the sequence level. DeepSTPP \cite{deepstpp} abandons absolute position encoding in favor of continuous sinusoidal position encoding based on normalized event time, enabling the attention mechanism to effectively address highly irregular time intervals in continuous time series. FEDformer \cite{fedformer} introduces frequency-enhanced attention by mapping temporal data to the frequency domain via Fourier or wavelet transforms, performing attention computations only on randomly retained frequency patterns, and then transforming back to the time domain to better capture global trends. iTransformer \cite{itransformer} embeds the entire global time series of each independent variable as a single token and computes self-attention across the variable dimension to capture complex correlations among multiple variables. HPformer \cite{hpformer} proposes a hierarchical propagation mechanism for temporal dependencies, which divides long sequences into several blocks and generates compressed key-value matrices through intra-block and inter-block attention, thereby efficiently capturing long-range temporal dependencies.

Regarding the causal attention mechanisms, CATT \cite{catt} first proposes causal attention by combining in-sample attention with cross-sample attention to reconstruct the attention score computation, effectively blocking spurious correlations in the data. GSAT \cite{gsat} injects randomness into attention weights to block task-irrelevant information and adaptively reduces attention randomness on label-relevant subgraphs, thereby removing spurious correlations. CAL \cite{cal} uses attention modules to estimate causal features and confounder features of the graph, respectively, performs random intervention by combining causal features with diverse confounder features, and thus eliminates confounding effects while extracting the causal skeleton. MULAN \cite{mulan} proposes a time series-aware attention mechanism. After extracting multimodal features through contrastive learning, it uses attention to dynamically evaluate the reliability weights of different modal time series data under the current fault, and then fuses them to generate a causal graph. E$^2$-CSTP \cite{e2cstp} introduces cross-modal attention and dual-branch causal intervention. The main branch processes spatiotemporal sequences, while the auxiliary branch uses cross-modal attention to dynamically block confounders and restore the true multimodal spatiotemporal causal relationships. CausalNET \cite{causalnet} is the first to apply causal attention mechanisms to non-i.i.d. topological event sequences, proposing a causal-aware Transformer-based event prediction module that uses a learnable causal graph to identify the causal types that have a direct impact on prediction. CausalFormer \cite{causalformer} proposes a multivariate causal attention mechanism, where Query and Key are generated based on sequence embeddings and Value uses the output of causal convolution, while introducing a learnable mask matrix in attention computation to adjust the sparsity of the causal graph.

Despite the significant progress of attention mechanisms in the above works, research targeting non-i.i.d. discrete event sequences remains limited. Specifically, existing methods often neglect prior knowledge such as network topology, lack domain adaptations for telecommunication networks, and fail to fully consider the complex causal relationships arising from concurrent events in the real world.

\section{Problem Formulation}
\label{sec:3}

\begin{figure*}
\centering
\includegraphics[width=0.98\linewidth]{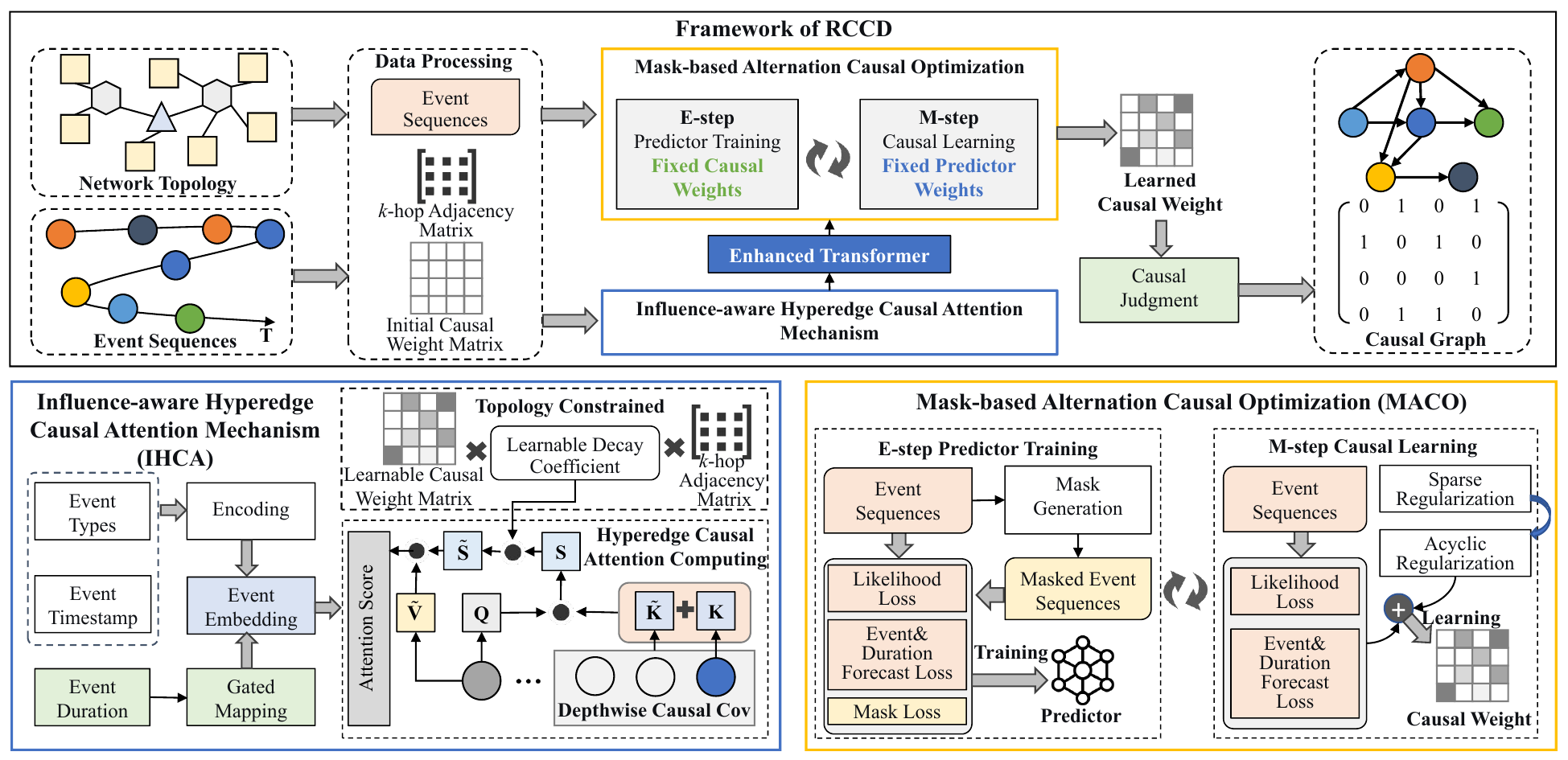}
\caption{Overview of the RCCD approach, which consists of two main components: Influence-aware Hyperedge Causal Attention (IHCA), which captures complex causal relationships by integrating event duration and concurrent effects while imposing topological constraints; Mask-based Alternation Causal Optimization (MACO), which enhances robustness to missing data through self-supervised reconstruction and optimizes the causal graph with sparsity and acyclicity regularization, ultimately generating an accurate directed acyclic causal graph. } 
\label{framework}
\end{figure*}

Given a telecommunication network, let the network topology be an undirected graph $\mathcal{G}_\mathrm{N} = (\mathcal{N}, \mathcal{E}_\mathcal{N})$, where $\mathcal{N}$ is the set of topological nodes and $\mathcal{E}_\mathcal{N}$ represents physical connections. To characterize the topological relationships among nodes, we define the $k$-hop neighborhood adjacency matrix $\mathrm{B}^k \in \{0,1\}^{|\mathcal{N}| \times |\mathcal{N}|}$, where $\mathrm{B}^k_{n_i, n_j} = 1$ if and only if node $n_j$ lies within the $k$-hop neighborhood of node $n_i$, and $0$ otherwise.
Let $\mathcal{V}$ be the set of event types, causal relationships among event types are represented by a directed acyclic graph $\mathcal{G}_\mathcal{V} = (\mathcal{V}, \mathcal{E}_\mathcal{V})$, where $v_i \rightarrow v_j \in \mathcal{E}_\mathcal{V}$ indicates that an event of type $v_i$ can cause an event of type $v_j$ to occur. We introduce a binary causal matrix $\mathrm{A} \in \{0,1\}^{|\mathcal{V}| \times |\mathcal{V}|}$, where $\mathrm{A}_{i,j} = 1$ if and only if there exists a causal edge $v_i \rightarrow v_j$, and $0$ otherwise.
Given an event sequence $\mathcal{X} = \{ x_i = (t_i, v_i, n_i, d_i) \}_{i=1}^\mathrm{L}$ containing $\mathrm{L}$ events, where $t_i \in \mathbb{R}^+$ is the timestamp, $v_i \in \mathcal{V}$ is the event type, $n_i \in \mathcal{N}$ is the topological node where the event occurs, and $d_i \ge 0$ is the duration of the event.

Our goal is to learn the causal relationships among event types from the observed event sequence $\mathcal{X}$ and the topological network $\mathcal{G}_N$, producing the final binary causal matrix $\mathrm{\textbf{A}}$.

\section{Methodology}
\label{sec:4}
In this section, we outline the overall architecture of the proposed method in Section \ref{sec:4.1}, clarifying the collaborative relationships among different components. We then provide detailed descriptions of the two innovative components in Sections \ref{sec:4.2} and \ref{sec:4.3}, respectively.

\subsection{Overview}
\label{sec:4.1}

In this paper, we propose a causal discovery method for telecommunication network alarm event sequences, aiming to robustly discover the directed acyclic causal graph among non-i.i.d. event sequences. As shown in Figure \ref{framework}, we first incorporate event duration, concurrency effects, and topological constraints into the attention computation of an enhanced transformer through a novel causal attention mechanism. Then, we adopt a mask-based multi-step optimization framework, enhance the predictor robustness to incomplete observations via a self-supervised reconstruction task, and jointly optimize the causal graph parameters and topology decay weights. Our method consists of two core components: an influence-aware hyperedge causal attention (IHCA) mechanism and a mask-based alternation causal optimization (MACO).

\textbf{IHCA:} To capture the complex causal relationships resulting from concurrent events, we first map event durations $d_i$ to embeddings and employ a gating mechanism to distinguish instantiated events, enabling the model to identify anomalous influences. Simultaneously, to address concurrent causality arising from multiple events, we design hyperedge causal attention, which aggregates consecutive events within a local time window into hyperedge representations via residual causal convolution. Furthermore, we jointly incorporate a learnable causal parameter matrix and a topological adjacency-based decay coefficient into the attention weights, ensuring that the model retains only event types permitted by topological distance.

\textbf{MACO:} To improve the robustness of causal learning to incomplete observations, we actively mask portions of historical events during predictor training in the E-step and add a reconstruction loss that forces the model to recover the masked event types using context. By compelling the predictor to break its reliance on the completeness of specific events, the predictor achieves robust representation to missing sequences. In the M-step, we fix the predictor parameters, optimize the causal parameters and topology decay coefficient using the unmasked original sequence, and introduce both a sparsity regularizer and a DAG-ness penalty to ensure that the learned causal graph is sparse and acyclic.

Overall, we first process the raw event sequences and network topology to obtain the event sequences, the $k$-hop adjacency matrix, and the initialized causal weight matrix. Subsequently, IHCA encodes event types, event timestamps, and event durations, computes attention scores via hyperedge causal convolution, and further multiplies attention scores with the causal weight matrix, $k$-hop adjacency matrix, and topology decay coefficients to obtain topologically constrained attention outputs. Furthermore, MACO integrates IHCA into the Transformer as an enhanced training model and alternately optimizes the predictor weights and causal graph weights through EM multi-step training. Finally, we make causal judgments based on the learned causal weights and predefined thresholds, outputting a binary causal graph.

\subsection{Influence-aware Hyperedge Causal Attention Mechanism}
\label{sec:4.2}

\subsubsection{Event Influence-Aware Encoding}

Existing methods \cite{s2gcsl,causalnet,tccd} typically leverage temporal dependencies in event sequences to achieve causal learning by predicting the next event. However, these methods ignore the impact of alarm duration. To capture the influence of different durations, we encode the event duration $d_i$ as a learnable feature in addition to the occurrence time encoding and event type encoding. Let the input sequence of the model be the matrix $\mathcal{X} \in \mathrm{R}^{\mathrm{L} \times \mathrm{d}}$, where $\mathrm{L}$ is the sequence length and $\mathrm{d}$ is the embedding dimension. For each alarm event $x_i = (t_i, v_i, n_i, d_i)$, we first map $d_i$ into a $\mathrm{d}$-dimensional embedding space through a nonlinear mapping network. Meanwhile, to avoid noise introduced by instantaneous events that have no duration, we apply gating via an indicator function. We define the duration embedding as

\begin{equation}
\mathrm{e}_i^\mathrm{U} = \Pi_{\{d_i > 0\}} \odot \mathcal{F}(d_i)
\end{equation}

where $\Pi$ is the indicator function, which takes the value 1 when $d_i > 0$ and 0 otherwise. $\mathcal{F}$ is a learnable nonlinear mapping function. $\odot$ denotes element-wise multiplication. Through the indicator function, the duration information is incorporated into the embedding $\mathrm{e}_i^\mathrm{U}$ only when the event is non-instantaneous. For an instantaneous event, its duration embedding is a zero vector. After obtaining the duration embedding $\mathrm{e}_i^\mathcal{U}$, we follow existing work to encode the event type and occurrence time using One-Hot vector mapping \cite{causalnet} and sinusoidal/cosine positional encoding \cite{thp}, respectively. Finally, the complete embedding $\mathrm{e}_i$ is formed by summing the three components of event type, occurrence time, and duration:

\begin{equation}
\mathrm{e}_i = \mathrm{e}_i^\mathcal{V} + \mathrm{e}_i^\mathcal{T} + \mathrm{e}_i^\mathcal{U}
\end{equation}

where $\mathrm{e}_i^\mathcal{V}$ is the event type embedding and $\mathrm{e}_i^\mathcal{T}$ is the occurrence time encoding. With this encoding scheme, the input sequence matrix $\mathcal{X}$ comprehensively encodes event type, occurrence time, and duration, providing fine-grained representations for downstream causal attention computation.

\subsubsection{Hyperedge Causal Attention Score}

The standard self-attention mechanism generates three vectors: Query (Q), Key (K), and Value (V), for each input embedding $\mathrm{e}_i$, allowing any two event embeddings to interact directly \cite{transformer}. However, the attention weight between events is determined solely by the dot product similarity between Query and Key, which fails to capture the complex causal relationships arising from multiple concurrent events. To address this issue, we design the hyperedge causal attention score, which dynamically aggregates consecutive events within a local time window into a hyperedge embedding via residual causal convolution. To focus on different feature patterns, we adopt multi-head attention, projecting the input into distinct representation subspaces. For each attention head $h$, we first map $\mathrm{e}_i$ into $\mathrm{Q}$, $\mathrm{K}$, and $\mathrm{V}$ through three different linear transformation matrices:

\begin{equation}
	\mathrm{Q} = \mathrm{e}_i \mathrm{W}^\mathrm{Q},\quad
	\mathrm{K} = \mathrm{e}_i \mathrm{W}^\mathrm{K},\quad
	\mathrm{V} = \mathrm{e}_i \mathrm{W}^\mathrm{V}
\end{equation}

where $\mathrm{W}^\mathrm{Q}, \mathrm{W}^\mathrm{K}, \mathrm{W}^\mathrm{V} \in \mathbb{R}^{d \times d_k}$ are weight matrices and $d_k$ is the dimension of each head. To model the concurrent effect of multiple events within a local event window, we apply depthwise causal convolution \cite{deepwise, tcn} to $\mathrm{K}$ and $\mathrm{V}$ respectively, and preserve the original information via residual connections. Let $\Psi(\cdot)$ denote the depthwise causal convolution. The enhanced $\tilde{\mathrm{K}}$ and $\tilde{\mathrm{V}}$ are given by

\begin{equation}
	\tilde{\mathrm{K}} = \mathrm{K} + \Psi(\mathrm{K}),\qquad
	\tilde{\mathrm{V}} = \mathrm{V} + \Psi(\mathrm{V})
\end{equation}

Specifically, the depthwise causal convolution $\Psi(\cdot)$ aggregates the current event and its two preceding events into a hypernode, enabling a single $\mathrm{K}$ or $\mathrm{V}$ vector to encode the joint features of multiple local events. Moreover, $\Psi(\cdot)$ applies one-dimensional causal convolution independently to each feature channel of $\mathrm{K}$ and $\mathrm{V}$, ensuring that at time step $i$, only positions before the event are accessible without leaking future information. We further employ residual connections to preserve the original single-event information. Consequently, in the attention computation, $\mathrm{Q}$ can directly interact with hypernodes, thereby capturing the complex causal relationship where multiple cause events jointly lead to a single effect. We compute the attention score using the enhanced $\tilde{\mathrm{K}}$ and $\tilde{\mathrm{V}}$:

\begin{equation}
	\mathrm{S} = \operatorname{softmax}\!\left( \frac{\mathrm{Q} (\tilde{\mathrm{K}})^{\top}}{\sqrt{d_k}} \right) 
\end{equation}

where $\mathrm{S} \in \mathbb{R}^{\mathrm{H} \times \mathrm{L} \times \mathrm{L}}$ is normalized by the softmax activation function, and the hyperedge causal attention transfers the aggregated historical information to the next layer with adaptive weights, allowing the model to capture both pairwise similarities and local cooperative effects. Therefore, the hyperedge causal attention score is more suitable than standard self-attention for discovering complex many-to-one causal relationships.

\subsubsection{Topology-Constrained Causal Attention Mechanism}

To inject network topology constraints and causal prior knowledge into attention, we further design a topology-constrained causal attention mechanism. Given that the current last known event is at position $i$, meaning the observed event is $e_i$, and the next event to predict is $e_{i+1}$, the raw attention score matrix computed by the hyperedge causal attention is $\mathrm{S} \in \mathbb{R}^{\mathrm{H} \times \mathrm{L} \times L}$, where $\mathrm{S}_{h,i,j}$ denotes the normalized attention weight for head $h$ with event $i$ as Query and event $j$ as Key. 

First, we construct a causal weight matrix $\mathrm{A}$ to ensure that direct influences occur only between types permitted by causality. Following \cite{causalnet}, let $\tilde{\mathrm{A}}=\omega(\theta) \in (0, 1)^{|\mathcal{V}| \times |\mathcal{V}|}$ be a learnable causal probability graph. A binary causal graph $\mathrm{A} \in \{0, 1\}^{|\mathcal{V}| \times |\mathcal{V}|}$ is sampled via Gumbel-Softmax function $\omega$ \cite{gumbelsoftmax}. Then, we construct a topology decay coefficient $\delta_{j,i+1}$ such that the influence between distant nodes is adaptively attenuated. Let the historical event $e_j$ occur at node $n_j$ and the event to be predicted $e_{i+1}$ occur at node $n_{i+1}$, the topology decay coefficient is defined as

\begin{equation}
\delta_{j,i+1} = \sum_{k=0}^Z \mathrm{B}^k_{n_j, n_{i+1}} \cdot \omega(\phi)
\end{equation}

where $\phi$ is a trainable topology decay weight, making the correlation lower for events that are farther apart in the topological distance \cite{graphattention}. Next, we introduce the causal weight matrix and the topology decay coefficient into the attention mechanism as prior constraints. For each attention head $h$, we multiply the raw attention weight element-wise with the causal weight and the topology decay to obtain the topology-constrained attention score:

\begin{equation}
\tilde{\mathrm{S}}_{h,i,j} = \mathrm{S}_{h,i,j} \cdot \tilde{\mathrm{A}}_{v_j, v_{i+1}} \cdot \delta_{j,i+1}, \quad \forall j \leq i
\end{equation}

We then compute the final output for that attention head by taking the weighted sum of the enhanced attention scores $\tilde{\mathrm{S}}$ and $\tilde{\mathrm{V}}$:

\begin{equation}
\text{Attention}(\mathrm{Q}, \tilde{\mathrm{K}}, \tilde{\mathrm{V}}) = \tilde{\mathrm{S}} \tilde{\mathrm{V}}
\end{equation}

Here, $\tilde{\mathrm{S}}$ has dimension $\mathrm{L} \times \mathrm{L}$ and $\tilde{\mathrm{V}}$ has dimension $\mathrm{L} \times d_k$. As a result, the attention mechanism retains only those event pairs that are both allowed by the causal graph and topologically close, while preserving temporal causality. 

As the core component of the transformer model, the proposed attention mechanism is computed independently in each attention head. The outputs from the $h$ heads are concatenated along the feature dimension and then linearly transformed to produce the multi-head attention result. Subsequently, the model stabilizes training through residual connections and layer normalization, and passes the result to a feedforward network for nonlinear transformation. Finally, the model predicts the next event in an autoregressive manner, allowing the causal graph and topology decay parameters to be continuously optimized via end-to-end gradients of sequence modeling, thereby achieving topology-constrained causal structure learning.

\subsection{Mask-based Alternation Causal Optimization}
\label{sec:4.3}

For causal discovery on event sequences, the existing EM-style alternating training framework first fixes the causal graph to train the prediction model, then fixes the prediction model to optimize the causal weights, alternating until convergence, and converts the discrete directed acyclic graph search into a differentiable continuous optimization \cite{thps, tnpar, causalnet}. However, the predictor in the E-step may learn incorrect temporal dependencies, causing the M-step causal graph optimization to generate numerous spurious edges and thus harming the robustness of causal discovery. To address this issue, we propose a mask-based alternating optimization framework. During predictor training, we actively mask a portion of historical events and force the model to reconstruct their attributes based on context, which compels the predictor to break its reliance on the completeness of specific events, making the model robust to missing noise.

\subsubsection{Mask Generation Strategy}

Let the input event sequence $\mathcal{X} = \{ x_i = (t_i, v_i, n_i, d_i) \}_{i=1}^\mathrm{L}$ contain $\mathrm{L}$ alarm events. Here, $\mathcal{T}$ is the occurrence time matrix, $\mathcal{V}$ the event type matrix, $\mathcal{N}$ the topological node matrix, and $\mathcal{U}$ the duration matrix. For an event sequence of length $\mathrm{L}$, we define a mask matrix $\mathrm{R} \in \{0,1\}^{\mathrm{B} \times \mathrm{L}}$, where $\mathrm{B}$ is the batch size and $1 \leq b \leq B$ indexes the $b$-th sequence in the batch. $\mathrm{R}_{b,i}=1$ indicates that position $(b,i)$ is masked. To avoid masking invalid or inappropriate positions, we impose the constraint that a position can be masked only when it satisfies both of the following conditions:

(i) Non-padding position ($v_{b,i} \neq 0$): the sequence may contain padding tokens whose event type is 0 and do not represent real events.

(ii) Non-prediction-target position ($i \neq \mathcal{N}_b$): the last valid position of each sequence is used as the prediction target during training. Masking this position would make it impossible to compute the prediction loss.

Here, $\mathcal{N}_b$ denotes the total number of events in the $b$-th sequence. Each valid position is independently masked with probability $p$. Let $\tilde{v}_{b,i}$ denote the event type after masking, defined as

\begin{equation}
	\tilde{v}_{b,i} = 
	\begin{cases}
		|\mathcal{V}|+1, & \mathrm{R}_{b,i}=1\\
		v_{b,i}, & \text{otherwise}
	\end{cases}
\end{equation}

where $|\mathcal{V}|+1$ is a dedicated mask token. By forcing the predictor to learn to reconstruct masked events, this strategy yields representations that are insensitive to partial observations, thereby mitigating the interference of missing data noise on causal graph learning. After masking the original event type matrix $\mathcal{V}$ with the mask matrix $\mathrm{R}$, we obtain the masked event type matrix $\tilde{\mathcal{V}}$.

\subsubsection{EM-Style Two-Stage Alternating Optimization Framework}

\textbf{E-Step Predictor Training.} In the first stage, we fix the current causal weight $\tilde{\mathrm{A}}$, the topology decay coefficient $\boldsymbol{\phi}$, and the $k$-hop Adjacency matrix $\mathrm{B}^k$. Taking the masked event type matrix $\tilde{\mathcal{V}}$, the occurrence time matrix $\mathcal{T}$, the topological node matrix $\mathcal{N}$, and the duration matrix $\mathcal{U}$ as input, the predictor outputs:

\begin{equation}
	v_j, t_j, u_j = \mathcal{P}\bigl(\tilde{\mathcal{V}},\; \mathcal{T},\; \mathcal{N},\; \mathcal{U},\; \mathrm{A},\; \boldsymbol{\phi},\; \mathrm{B}^k\bigr).
\end{equation}

Here, $\mathcal{P}$ is the predictor, a transformer implemented with the influence-aware hyperedge causal attention mechanism. $v_j$ is the predicted next event type, $t_j$ the predicted next time interval, and $u_j$ the reconstruction prediction for masked positions. To achieve efficient training of the predictor, we employ four loss components: event intensity likelihood loss, time interval prediction loss, event type prediction loss, and a specially designed mask reconstruction loss that acts as a regularizer to ensure robust prediction performance. To guide the predictor in learning the true temporal dependence patterns of event sequences, we first compute the event intensity likelihood loss based on the intensity function of the Hawkes process \cite{isotonichp}:

\begin{equation}
	\mathcal{L}_{\text{H}} = -\sum_{i=1}^{\mathrm{L}-1} \left( \log \lambda_{v_{i+1}}(t_{i+1} \mid \mathrm{H}_{t_i}) - \int_{t_i}^{t_{i+1}} \lambda_{v_{i+1}}(t \mid \mathrm{H}_{t_i}) dt \right)
\end{equation}

where $\mathrm{H}_{t_i}$ denotes the set of historical events up to time $t_i$, and the intensity function $\lambda_{v}(t\mid \mathrm{H}_{t_i})$ \cite{sahp} represents the probability of an event of type $v$ occurring at time $t$ given the history $\mathrm{H}$. Notably, the loss $\mathcal{L}_{\text{H}}$ consists of a log term and an integral term. Specifically, the log term maximizes the intensity at the actual event occurrence time, while the integral term penalizes overly high predicted intensity for that event type during intervals with no events. Although the event intensity prediction loss can predict events from the perspective of occurrence intensity, it is weak at discriminating event types themselves and lacks precise prediction of time intervals. Therefore, we further design an event type prediction loss and a time interval prediction loss to improve model performance:

\begin{equation}
	\mathcal{L}_{\text{V}} = -\sum_{i=1}^{{\text{L}}-1} \log \hat{v}_{i+1}[v_{i+1}],  \\
	\mathcal{L}_{\text{T}} = \sum_{i=1}^{\text{L}-1} \bigl( \hat{\tau}_{i+1} - (t_{i+1} - t_i) \bigr)^2
\end{equation}

In particular, the event type prediction loss $\mathcal{L}_{\text{V}}$ adopts the cross-entropy form, minimizing the negative log probability of the true type, enabling the model to accurately discriminate the category of the next event. Complementing $\mathcal{L}_{\text{V}}$, the time interval prediction loss $\mathcal{L}_{\text{T}}$ uses mean squared error, directly penalizing the deviation between the predicted interval and the true interval, allowing the model to precisely locate event occurrence times. Both losses assume that the input sequence is fully observed and focus only on the prediction performance of future events. However, event missing in telecommunication networks may cause the predictor to learn incorrect temporal relationships. To address this, we introduce a mask reconstruction loss to enhance robustness through a self-supervised learning task:

\begin{equation}
	\mathcal{L}_{\text{R}} = -\frac{1}{\|\mathrm{R}\|_1} \sum_{b=1}^\text{B} \sum_{i=1}^\text{L} \mathrm{R}_{b,i} \cdot \log \hat{u}_{b,i} \cdot v_{b,i}
\end{equation}

Here, $\mathrm{R}_{b,i}=1$ indicates a masked position, and $\hat{u}_{b,i}[v_{b,i}]$ is the predicted probability of the reconstruction head for the true type $v_{b,i}$. By minimizing the negative log probability at masked positions, the model learns to complete missing event using contextual information, effectively mitigating the interference of event missing noise in telecommunication networks. Based on the event intensity likelihood loss, event type prediction loss, time interval prediction loss, and mask reconstruction loss, the total loss function in the prediction stage is:

\begin{equation}
	\mathcal{L}_\mathcal{P} = \mathcal{L}_{\text{H}} + \mathcal{L}_{\text{T}} + \mathcal{L}_{\text{V}} + \mathcal{L}_{\text{R}}
\end{equation}

By minimizing $\mathcal{L}_\mathcal{P}$, the predictor is jointly optimized across the four tasks, thereby learning representations robust to missing data and significantly improving the stability and accuracy of causal discovery in telecommunication network scenarios.

\textbf{M-Step Causal Learning.} In the second stage, we fix all parameters of the predictor $\mathcal{P}$ and use the unmasked original sequence to optimize the causal weights $\tilde{\mathrm{A}}$ and the topology decay coefficient $\boldsymbol{\phi}$. We have summarized the loss function for this stage as follows:

\begin{equation}
	\mathcal{L}_{\mathcal{M}} = \bigl(\mathcal{L}_{\text{H}} + \mathcal{L}_{\text{T}} + \mathcal{L}_{\text{V}} \bigr) + \sigma_1 \|\omega(\boldsymbol{\tilde{\mathrm{A}}})\|_1 + \sigma_2  h\bigl(\omega(\boldsymbol{\tilde{\mathrm{A}}})\bigr)
\end{equation}

where $\sigma_1, \sigma_2$ are tunable hyperparameters, $\omega$ is an activation function, and $\|\cdot\|_1$ is the $\ell_1$ norm. Here, the sparsity regularizer $\|\omega(\boldsymbol{\tilde{\mathrm{A}}})\|_1$ encourages the causal graph to remain sparse and avoid redundant spurious causal edges, and the DAG-ness penalty $h(\cdot)$ comes from NOTEARS \cite{notears}, satisfying $h(\mathrm{A}) = 0$ if and only if $\tilde{\mathrm{A}}$ is a directed acyclic graph. Using $\mathcal{L}_{\mathcal{M}}$, we can directly optimize $\boldsymbol{\tilde{\mathrm{A}}}$ via gradient descent, thereby efficiently searching the causal graph space. By alternately executing the E-step predictor training and the M-step causal graph optimization, the predictor and the causal graph reinforce each other, eventually converging to a sparse and acyclic causal graph.

\subsubsection{Causal Judgment}

After EM training, we determine whether a causal edge exists between event types based on the learned causal weight matrix $\omega(\boldsymbol{\tilde{\mathrm{A}}})$. Specifically, for each pair of event types, if the corresponding causal probability $\omega(\boldsymbol{\tilde{\mathrm{A}}}_{i,j})$ exceeds a preset threshold $\rho$, we determine that a causal edge exists from $v_i$ to $v_j$; otherwise, we determine that no such edge exists. Formally, this decision rule can be expressed as:

\begin{equation}
A_{i,j} =
\begin{cases} 
	1, & \text{if } \omega(\boldsymbol{\tilde{\mathrm{A}}}_{i,j}) > \rho \\
	0, & \text{otherwise}
\end{cases}
\end{equation}

It is worth noting that during training, we introduce the topology decay coefficient $\phi$ into the attention weight computation, enabling the model to jointly learn the causal probability and the topological structure, thereby improving the quality of the learned causal graph.

\section{Experiment}
\label{sec:6}

To evaluate the effectiveness of RCCD for causal discovery from event sequences, we design five sets of experiments. First, we compare the causal graph learning performance of RCCD with existing baselines on stationary datasets. Second, we test the robustness of RCCD under incomplete observations. Third, we validate the impact of key hyperparameters on causal discovery results. Fourth, we conduct ablation studies by removing each core component to quantify its contribution to overall performance. Finally, we conduct a case study using a real-world dataset and visually present the causal graph discovered by RCCD.

\subsection{Dataset}

\textbf{Synthetic Datasets:}
To evaluate causal discovery performance, we generate multi-type event sequences using the gCastle API \cite{gcastle}. gCastle allows control over the number of event types, number of physical devices, causal strength, and number of events, and is widely used for benchmarking causal discovery methods. For each synthetic dataset, we first generate a ground-truth Granger causal graph and an undirected topology graph among the nodes. We then generate root events via a Poisson process and propagate events according to the causal structure and topology graph with randomly sampled excitation parameters. Following existing work \cite{causalnet, s2gcsl, tccd}, we construct two synthetic datasets. Dataset 15V-30N-Synthetic contains 15 event types and 30 nodes. Dataset 20V-40N-Synthetic contains 20 event types and 40 nodes. The baseline intensity $\mu \in [3 \times 10^{-5}, 5 \times 10^{-5}]$, the excitation intensity $\alpha \in [0.02, 0.03]$, the generation period is 2 days, and the sampling interval $\Delta = 1$.

\textbf{Metropolitan Cellular Network Alarm Dataset:}
To verify the generalization ability of RCCD in real-world scenarios, we adopt an event sequence dataset jointly released by Huawei and Peking University \cite{s2gcsl}. Dataset 24V-439N-Microwave contains 24 event types, 439 nodes, and 64,599 event records. Dataset 25V-474N-Microwave contains 25 event types, 474 nodes, and 48,573 event records. All data are generated from real networks, where alarm propagation paths reflect the diffusion process of faults incidents among devices, providing ground truth causal graphs, device-level network topology, and complete alarm event sequences. Due to the characteristics of the logging system, timestamps have specific discretization intervals. With differing scales, event sparsity, and event duration distributions, these two datasets enable a comprehensive evaluation of RCCD’s performance in complex real-world environments.

\subsection{Baselines}
We comprehensively compare the RCCD method with SOTA causal discovery methods for event sequences. PC (2000) \cite{pc} is a classic causal discovery algorithm based on conditional independence tests, constructing a causal graph by progressively removing edges. GraNDAG (2020) \cite{grandag} parameterizes nonlinear relationships among variables using neural networks and learns a directed acyclic graph under continuous optimization constraints. CORL (2021) \cite{corl} combines reinforcement learning with causal structure search, guiding the graph construction process through a reward mechanism. THPs (2022) \cite{thps} first incorporate network topology into the Hawkes process, explicitly modeling the diffusion process of faults among devices via graph convolution. TNPAR (2024) \cite{tnpar} introduces neural point processes on top of THPs, enhancing the representation capability for complex alarm sequences. S$^2$GCSL (2024) \cite{s2gcsl} employs a linear kernel Hawkes process and stochastic gradient descent to efficiently optimize the likelihood function, combining L$_1$ regularization and topological information to learn a sparse causal graph. CausalNET (2024) \cite{causalnet} introduces trainable causal weights and a topology decay matrix into the Transformer, optimizes the causal graph via Gumbel-Softmax, and alternately trains the prediction module and the graph structure. TCCD (2025) \cite{tccd} introduces a temporal decay term on top of S$^2$GCSL and integrates Monte Carlo tree search with continuous gradient optimization for both global and local search. Among these methods, PC, GraNDAG, and CORL do not use network topology prior knowledge, whereas THPs, TNPAR, S$^2$GCSL, CausalNET, and TCCD are advanced methods that incorporate network topology prior knowledge. We also compare three advanced large language models: DeepSeek-v3.2 \cite{deepseek}, Gemini-v3.0, and ChatGPT-v5.1 \cite{gpt51}.

\subsection{Evaluation Criteria}
Given the binary causal graph discovered by each method, we employ four standard metrics to evaluate causal learning performance: Precision, Recall, F1 Score, and Structural Hamming Distance. In computing these metrics, TP denotes that a causal edge exists in the ground truth causal graph and is correctly predicted by the causal discovery method. FN denotes that a causal edge exists in the ground truth graph but is not predicted by the method. TN denotes that no causal edge exists in the ground truth graph and the method also does not predict it. FP denotes that no causal edge exists in the ground truth graph but the method incorrectly predicts one. \textbf{Precision (PRE)} measures the accuracy of predicted edges: Precision = TP / (TP+FP). \textbf{Recall (REC)} measures the coverage of true edges: Recall = TP / (TP+FN).\textbf{ F1 Score (F1)} is the harmonic mean of Precision and Recall, providing a comprehensive evaluation of prediction performance: F1 = 2 * (Precision * Recall) / (Precision + Recall). \textbf{Structural Hamming Distance (SHD)} computes the minimum number of edge modifications (insertions, deletions, or reversals) required to transform the predicted causal graph into the ground truth causal graph, directly quantifying the overall structural difference. For Precision, Recall, and F1, higher values indicate better performance, while for SHD, lower values indicate better performance.

\subsection{Implementation and Environment}

For the proposed causal discovery method, the hyperparameter settings are as follows. For the 24V-439N-Microwave dataset, the number of topological neighbor hops is set to 1, while for the other datasets it is set to 2. Following the existing work \cite{causalnet}, we configure the Transformer encoder to have 4 layers, 4 attention heads, an embedding dimension of 512, a feedforward dimension of 1024, and a dropout rate of 0.1, and use Adam as the optimizer. Simultaneously, we set the initial temperature of the Gumbel-Softmax to 1.0, the final temperature to 0.1, the batch size to 512, and the maximum number of training iterations to 200. Experiments are conducted on a workstation equipped with an Intel® Xeon® Gold 6326 CPU, 503 GB RAM, and 8 NVIDIA L40 GPUs. The software environment consists of Ubuntu 22.04, CUDA 12.6, Python 3.9.21, and PyTorch 2.6.0.

\subsection{Performance Comparison on Stable Datasets}
\label{sec:stable}

\begin{table*}[ht]
\newcommand{\tabincell}[2]{\begin{tabular}{@{}#1@{}}#2\end{tabular}}
\centering
\begin{threeparttable}
\setlength\tabcolsep{4.0pt}
\caption{Performance comparison on Stable datasets}
\begin{tabular}{ccccccccccccccccc}
\toprule  
\multirow{2}*{Methods}& \multicolumn{4}{c}{15V-30N-Synthetic}& \multicolumn{4}{c}{20V-40N-Synthetic} & \multicolumn{4}{c}{24V-439N-Microwave}& \multicolumn{4}{c}{25V-474N-Microwave}\\
\cline{2-17}~&PRE & REC & F1 & SHD& PRE & REC & F1 & SHD& PRE & REC & F1 & SHD& PRE & REC & F1 & SHD\\
\midrule
PC&	0.0141&	0.1111&	0.0250&	77&	0.0451&	0.3529&	0.08&	136&	0.3243&	0.2628&	0.2903&	142&	0.2821&	0.223&	0.2491&	162\\
GraNDAG&	0.1667&	0.3333&	0.2222&	21&	0.0500&	0.0588&	0.0541&	34&	0.2500&	0.0438&	0.0745&	144&	0.4667&	0.0473&	0.0859&	143\\
CORL&	0.1667&	0.2222&	0.1905&	17&	0.1429&	0.1765&	0.1579&	32&	0.1176&	0.0146&	0.0260&	140&	0.2000&	0.0338&	0.0578&	158\\
\bottomrule
THPs&	0.4737&	1.0000&	0.6429&	10&	0.8000&	0.9412&	0.8649&	5&	0.6000&	0.0657&	0.1184&	129&	0.6000&	0.0811&	0.1429&	136\\
TNPAR&	0.1000&	0.2222&	0.1379&	25&	0.0630&	0.4706&	0.1111&	124&	0.2524&	0.9781&	0.4012&	142&	0.2462&	0.9932&	0.3946&	153\\
S$^2$GCSL&	0.3077&	0.8889&	0.4571&	12&0.4545&	0.8824&	0.6000	&19&	0.2407&	0.9489&	0.3840&	146&	0.2467&	1.0000&	0.3957	&152\\
CausalNet&	0.5385&	0.7778&	0.6364&	6	&0.5417	&0.7647&	0.6341&	15	&0.4245	&0.5547&	0.4810	&130&	0.4054&	0.5067&	0.4504&	152\\
TCCD&	0.4286&	0.5000	&0.4615&	21&	0.1429&	0.5294&	0.2250&	59&	0.2394&	0.9489&	0.3824&	146&	0.2467&	1.0000&0.3957&	152\\
\textbf{Proposed}&	\textbf{0.8000}&	\textbf{0.8889}&	\textbf{0.8421}&	\textbf{3}&	\textbf{0.9333}&	\textbf{0.8235}&	\textbf{0.8750}&	\textbf{4}&	0\textbf{.4130}&	\textbf{0.7445}&	\textbf{0.5313}&	\textbf{135}	&\textbf{0.3681}&	\textbf{0.7635}&	\textbf{0.4967}&	\textbf{158}\\
\bottomrule
\end{tabular}
\label{table1}
\end{threeparttable}
\end{table*}

\begin{figure*}[!htb]
	\centering
	\scriptsize
	\subfloat[15V-30N-Synthetic (F1$\uparrow$)]{
		\includegraphics[width=0.24\linewidth]{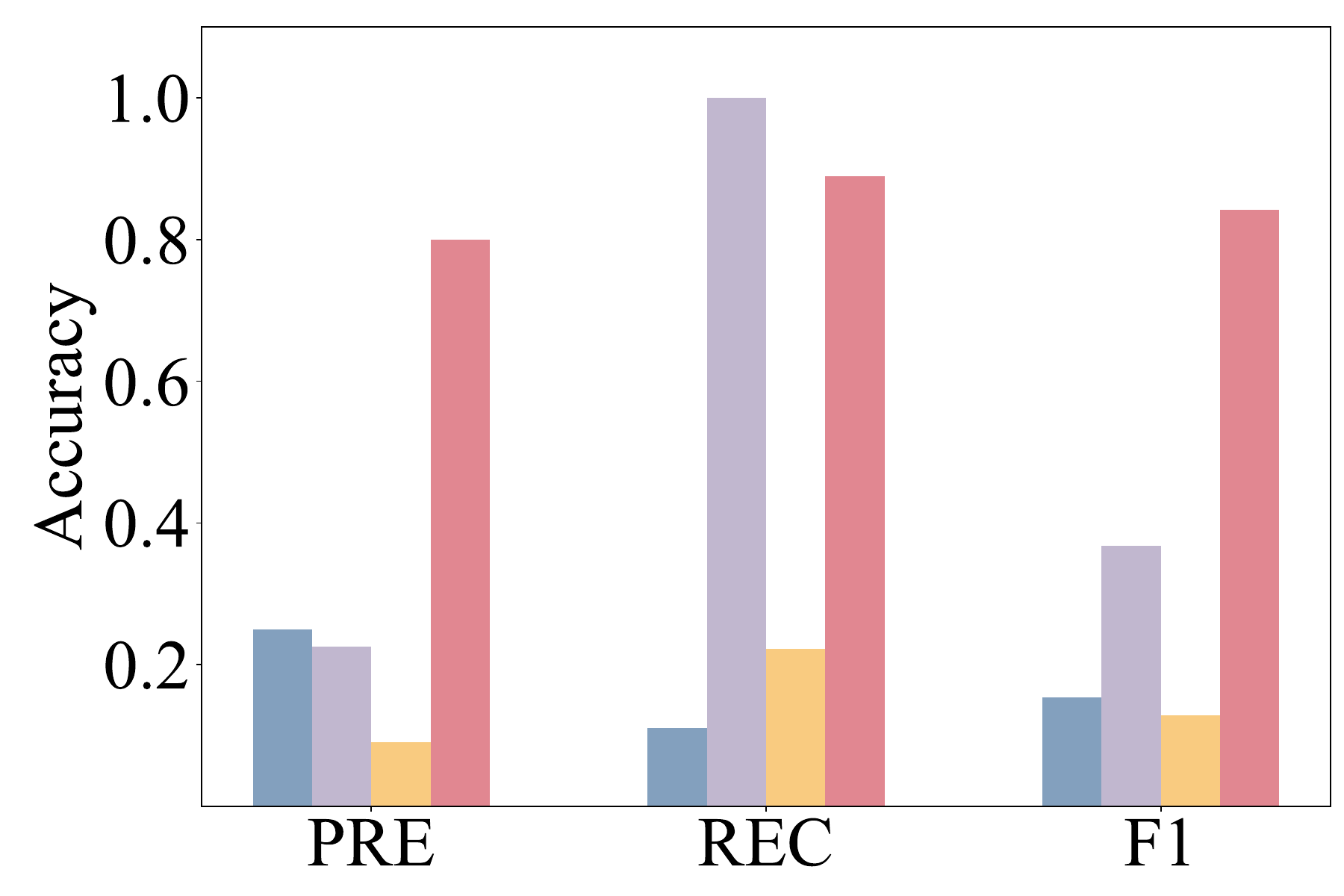}
	}
    \hfill
	\scriptsize
	\subfloat[20V-40N-Synthetic (F1$\uparrow$)]{
		\includegraphics[width=0.24\linewidth]{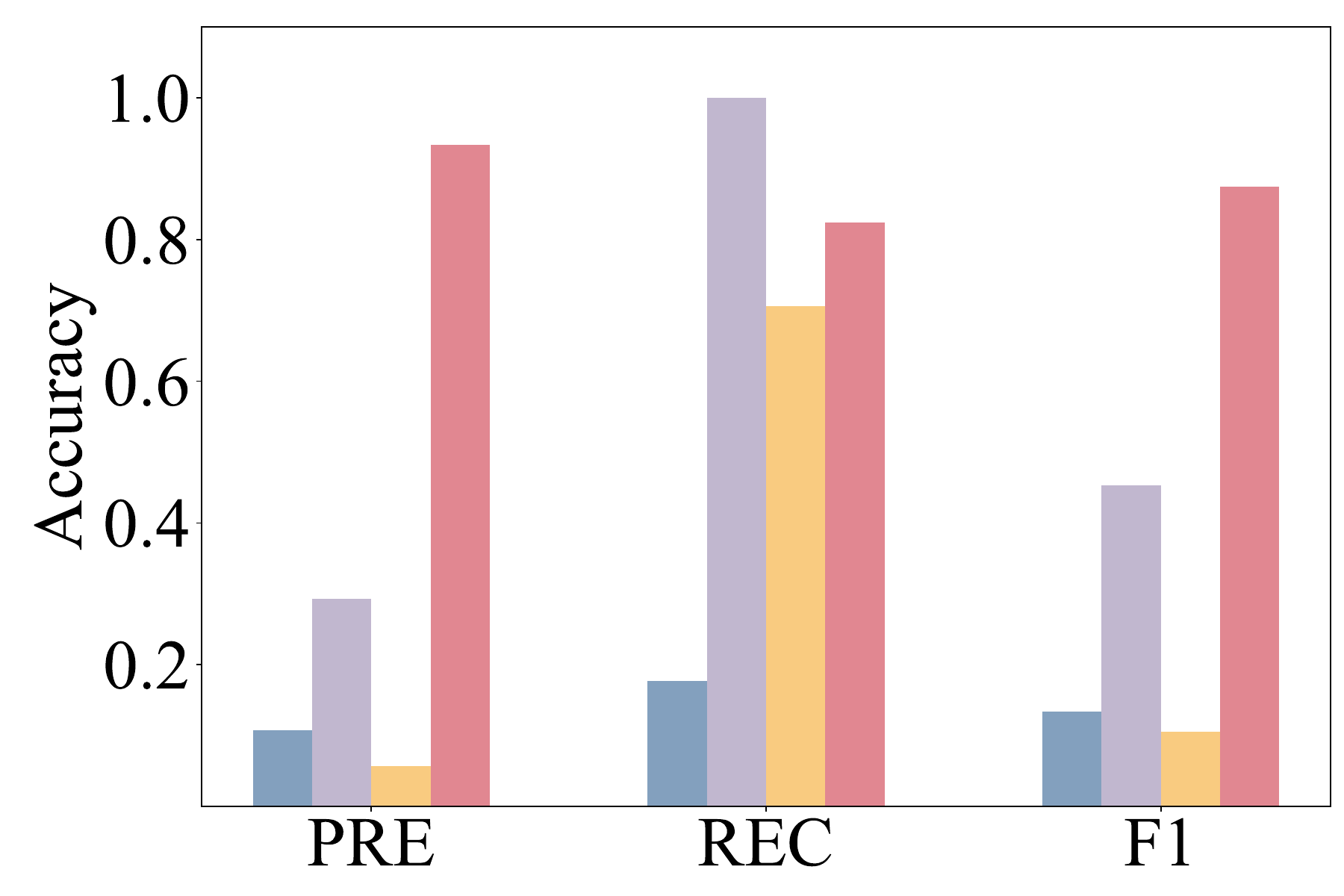}
	}
	\hfill
	\scriptsize
	\subfloat[24V-439N-Microwave (F1$\uparrow$)]{
		\includegraphics[width=0.24\linewidth]{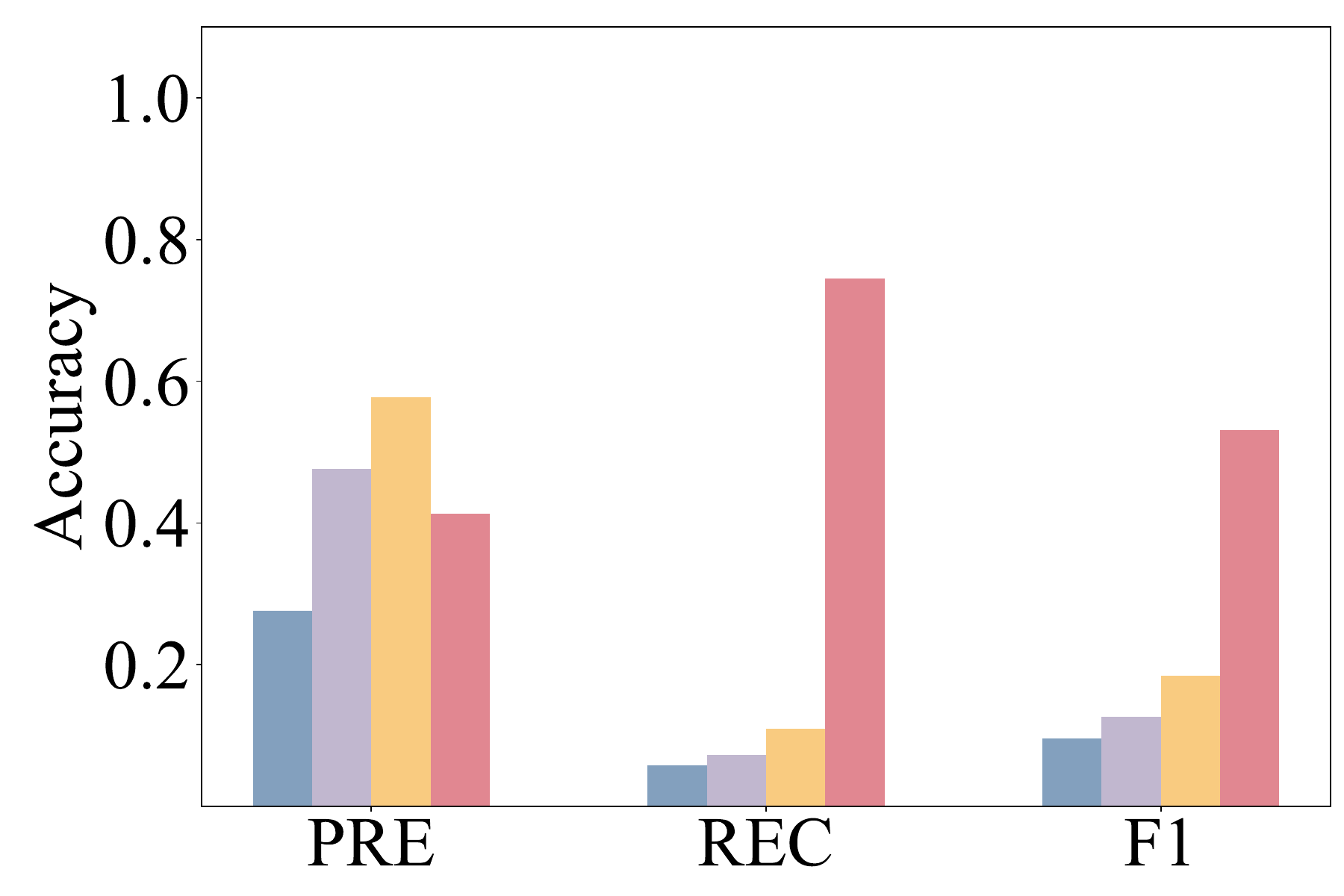}
	}
	\hfill
	\scriptsize
	\subfloat[25V-474N-Microwave (F1$\uparrow$)]{
		\includegraphics[width=0.24\linewidth]{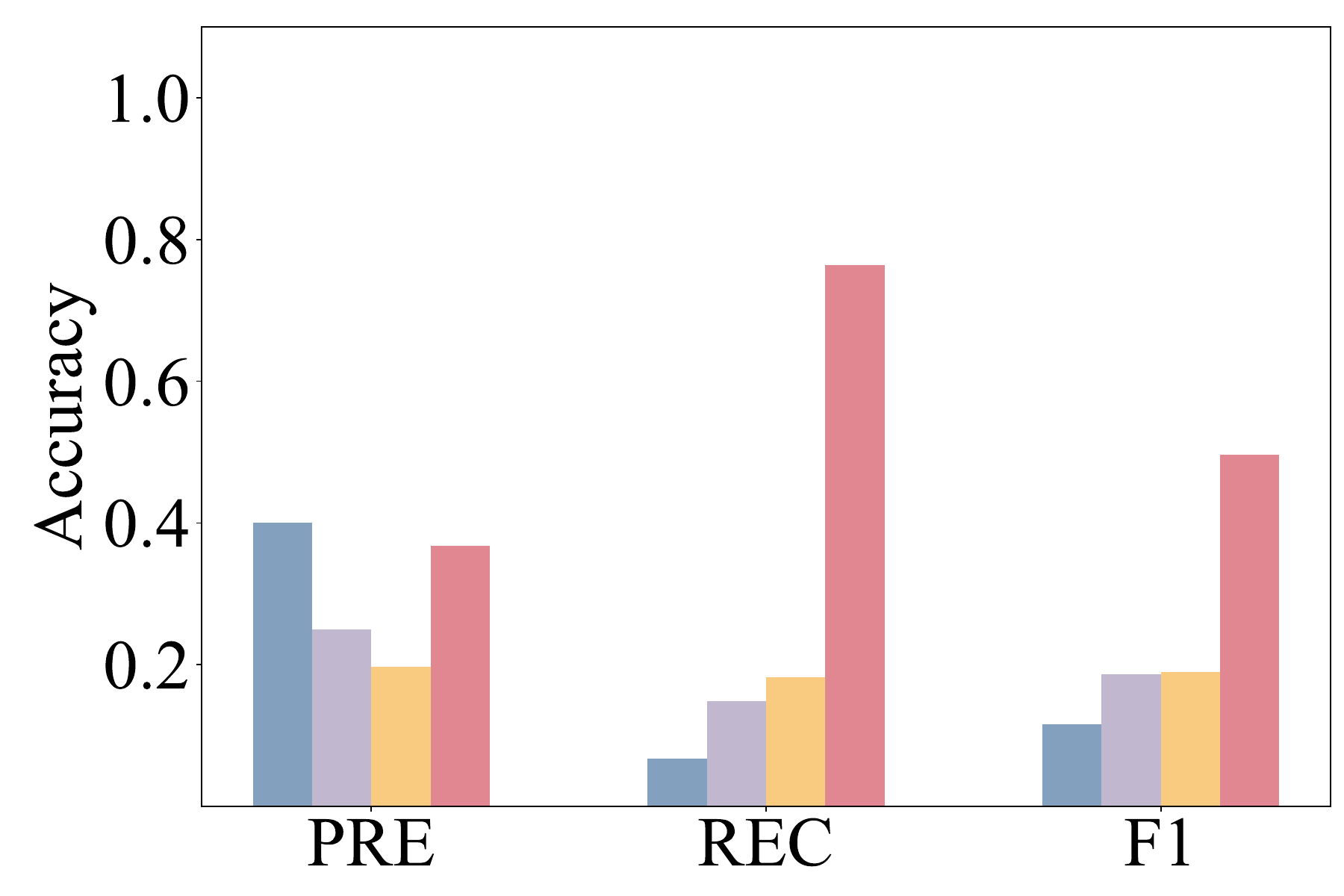}
	}
    
	\centering
	\scriptsize
	\subfloat[15V-30N-Synthetic (SHD$\downarrow$)]{
		\includegraphics[width=0.24\linewidth]{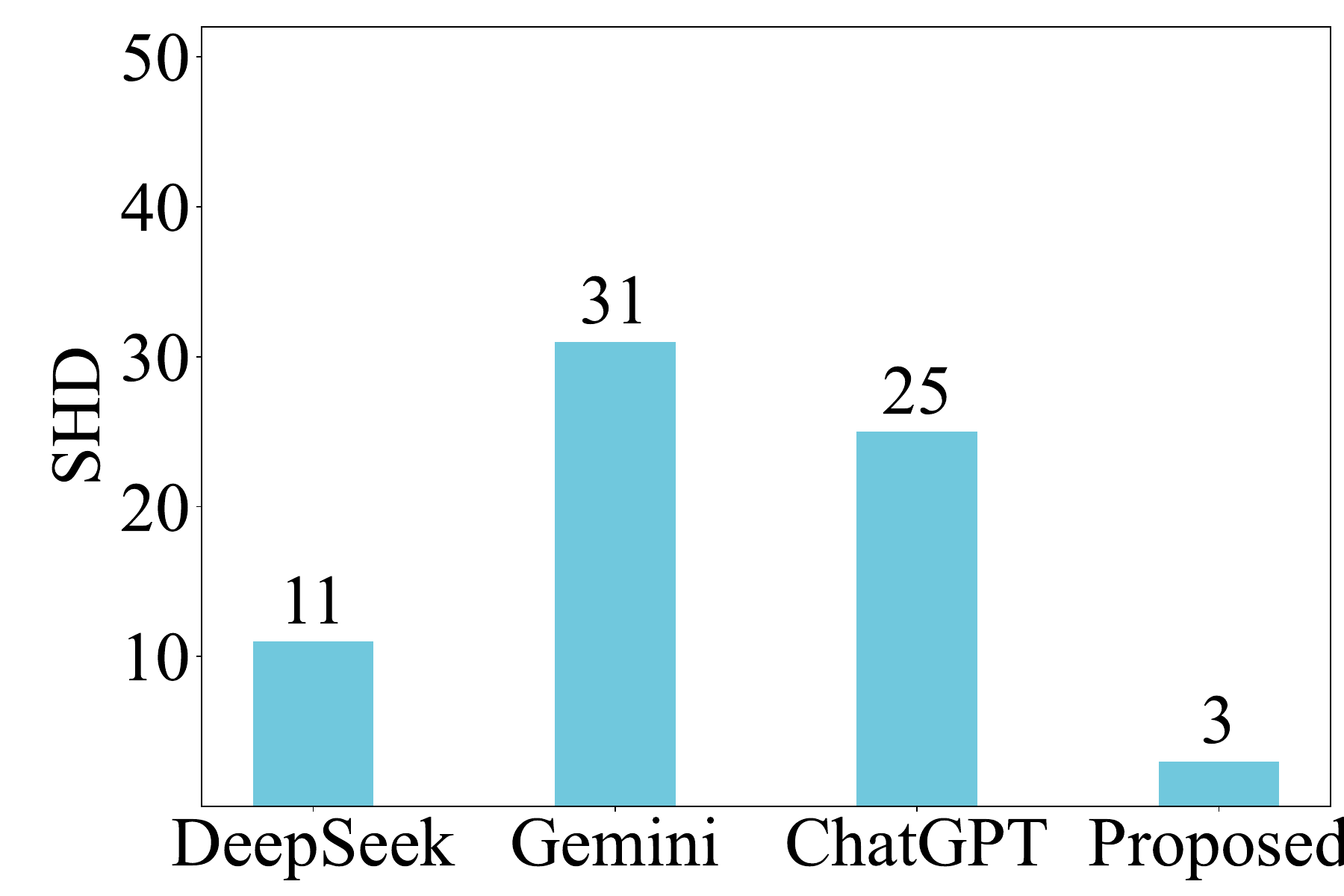}
	}
	\hfill
	\subfloat[20V-40N-Synthetic (SHD$\downarrow$)]{
		\includegraphics[width=0.24\linewidth]{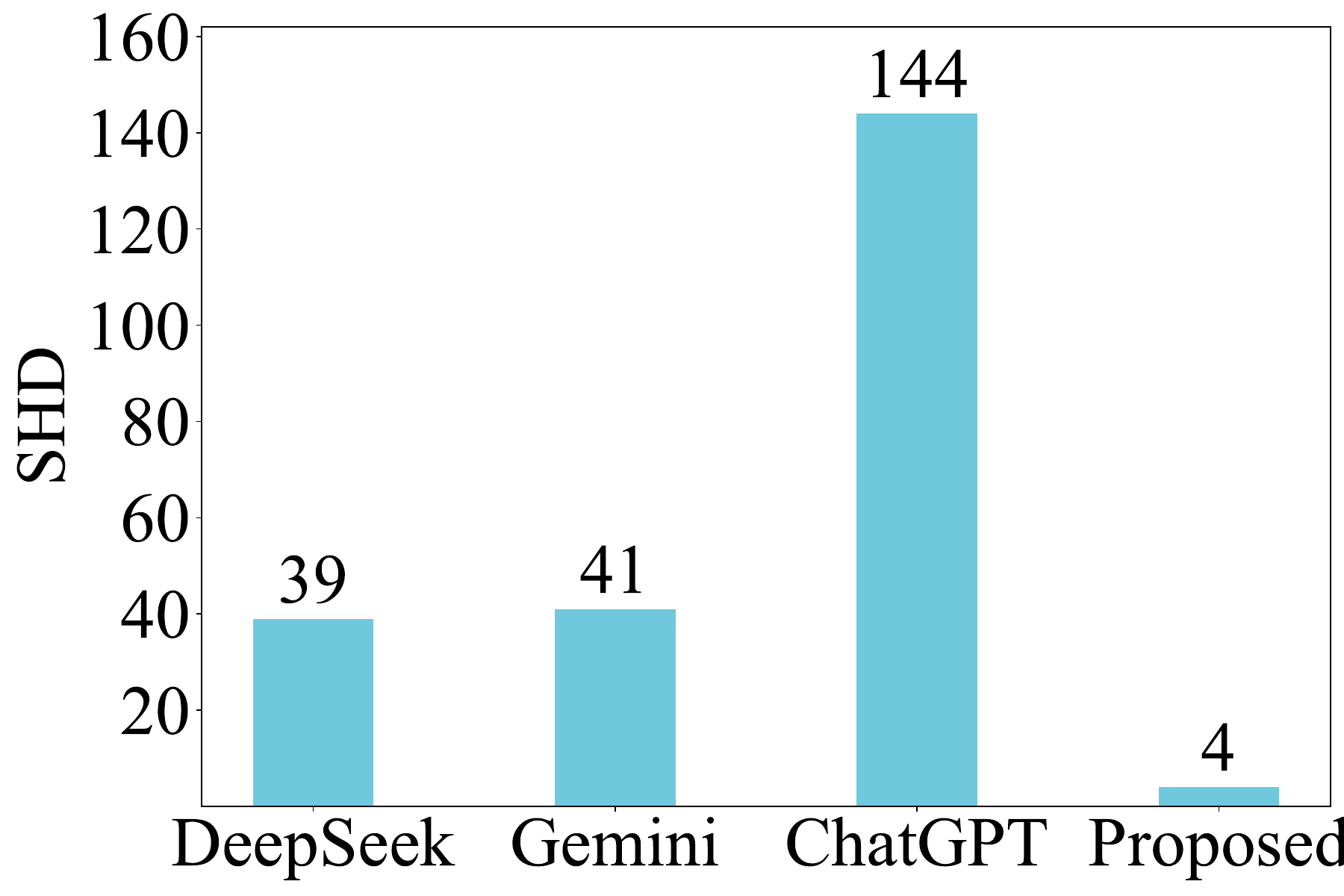}
	}
	\hfill
	\scriptsize
	\subfloat[24V-439N-Microwave (SHD$\downarrow$)]{
		\includegraphics[width=0.24\linewidth]{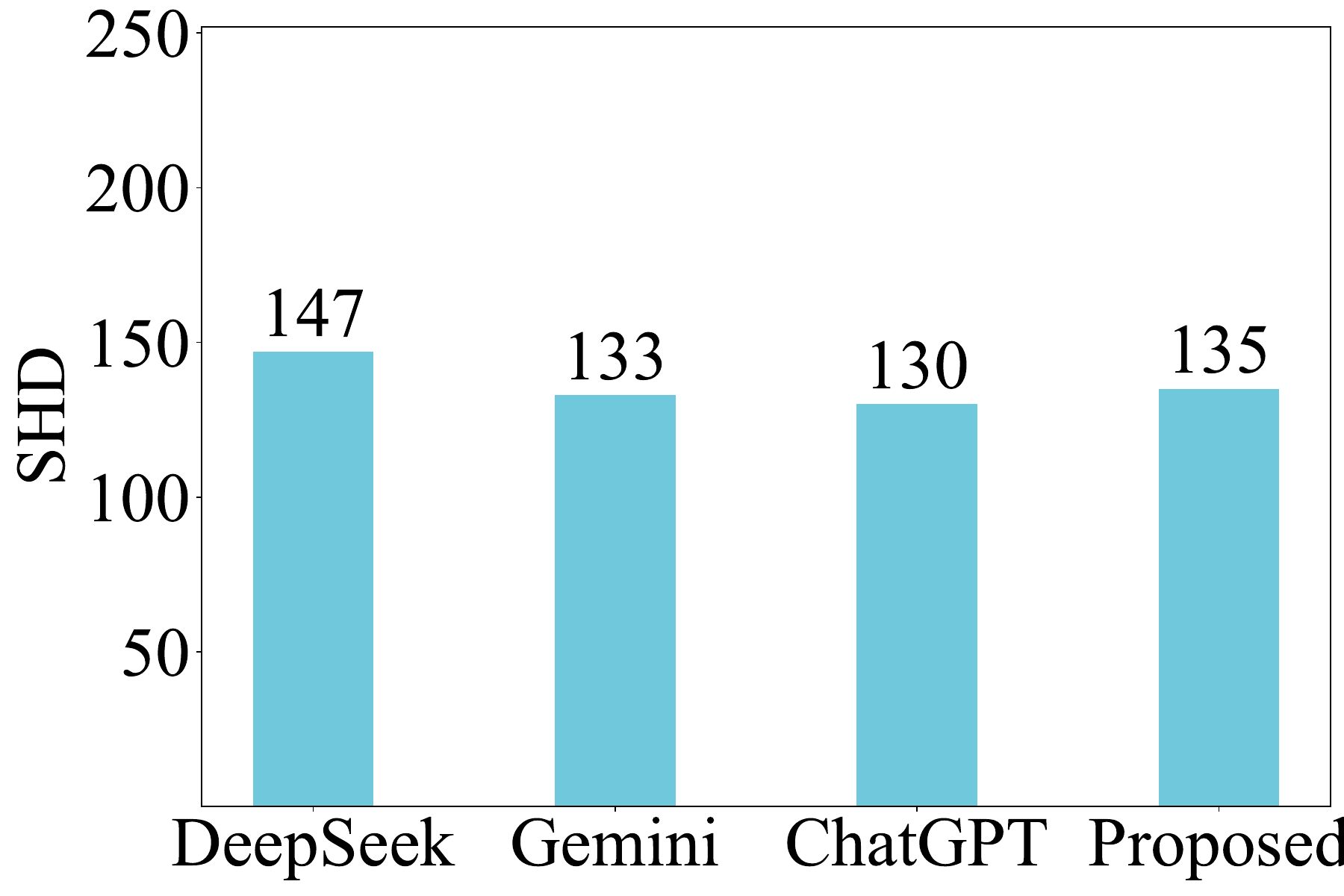}
	}
	\hfill
	\scriptsize
	\subfloat[25V-474N-Microwave (SHD$\downarrow$)]{
		\includegraphics[width=0.24\linewidth]{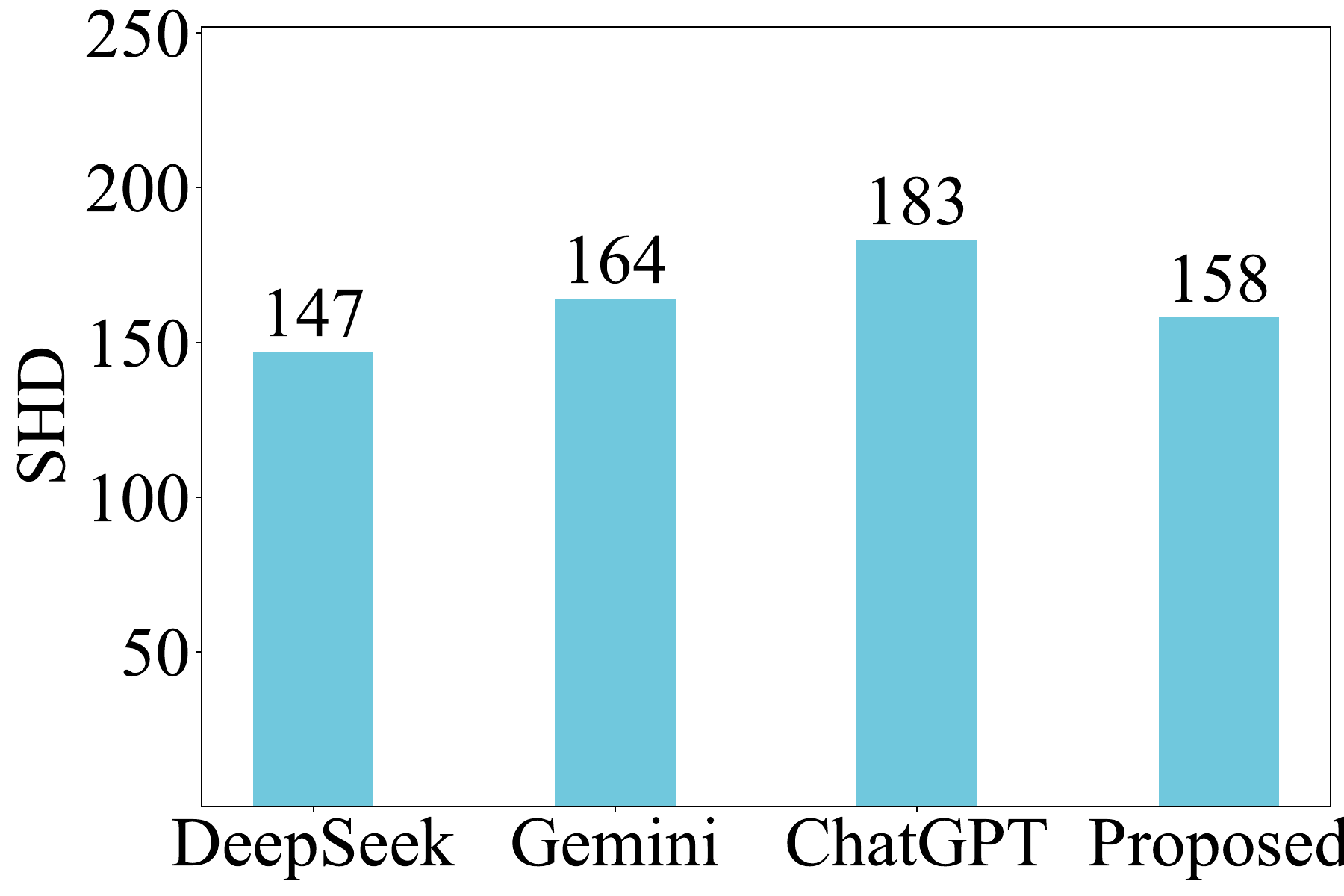}
	}
    
    \centering
    \scriptsize
	\subfloat{
	\includegraphics[width=0.4\linewidth]{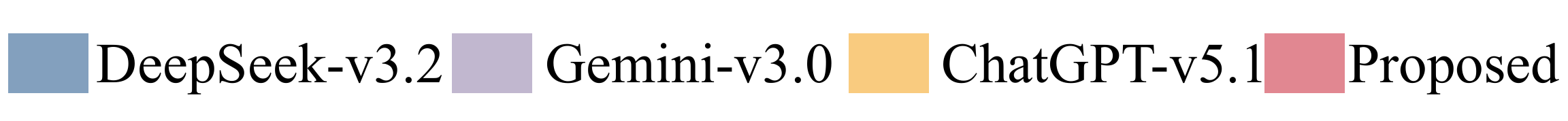}
    }
\caption{Performance Comparison of Advanced Large Language Models on Synthetic Datasets.}
\label{env:llm}
\end{figure*}

On Stable datasets, we comprehensively evaluate causal structure learning performance by comparing the F1 Score and SHD of the proposed method against various baselines. Overall, our method achieves the best or second-best F1 Score on all datasets and obtains the lowest SHD on the synthetic datasets, validating its superiority in causal discovery from stationary event sequences.

As shown in Table \ref{table1}, RCCD performs exceptionally well on both synthetic datasets in terms of F1 Score. On the 15V-30N-Synthetic dataset, RCCD achieves an F1 of 0.8421, much higher than the 0.6364 of CausalNET, the 0.6429 of THPs, and the 0.4615 of TCCD. On the 20V-40N-Synthetic dataset, our F1 is 0.8750, again significantly outperforming the 0.6341 of CausalNET and the 0.8649 of THPs. Notably, THPs achieves a recall close to 1.0 on this dataset, but its precision is only 0.8, whereas RCCD strikes a better balance between precision and recall. On the real-world datasets, RCCD achieves an F1 of 0.5313 on 24V-439N-Microwave, higher than the 0.4810 of CausalNET and the 0.1184 of THPs. Meanwhile, RCCD attains an F1 of 0.4967 on the 25V-474N-Microwave dataset, again leading over other baselines. In contrast, although TNPAR and S$^2$GCSL have very high recall, their low precision leads to unsatisfactory F1 Scores.

Regarding the Structural Hamming Distance, RCCD shows clear advantages on the synthetic datasets. On the 15V-30N-Synthetic dataset, the SHD is only 3, far lower than the 6 of CausalNET, the 10 of THPs, and the 21 of TCCD. On the 20V-40N-Synthetic dataset, our SHD is 4, also better than the 15 of CausalNET and the 5 of THPs, indicating that the causal graph discovered by RCCD is highly similar to the ground truth graph in terms of edge sets. However, on the real-world datasets, our SHD is 135 on the 24V-439N-Microwave dataset, slightly higher than the 130 of CausalNET. Nevertheless, RCCD still achieves the highest F1 Score, suggesting that the causal edges it predicts are more accurate, and the higher SHD mainly results from the addition of some weak causal edges rather than the omission of critical ones.

As illustrated in Figure \ref{env:llm}, compared with advanced large language models, RCCD significantly outperforms all three Large Language Models in F1 Score on every dataset. On the 15V-30N-synthetic dataset, our F1 of 0.8421 far exceeds the 0.3673 of Gemini, the 0.1538 of DeepSeek, and the 0.1290 of ChatGPT. On 20V-40N, our F1 of 0.8750 similarly surpasses the 0.4533 of Gemini, the 0.1333 of DeepSeek, and the 0.1057 of ChatGPT. On the real-world datasets, our F1 of 0.5313 on the 24V-439N-Microwave dataset is much higher than the 0.1840 of ChatGPT, the 0.1266 of Gemini, and the 0.0964 of DeepSeek. In terms of SHD, RCCD achieves an SHD of 3 on 15V-30N, much lower than the 31 of Gemini, the 25 of ChatGPT, and the 21 of DeepSeek. On the real-world datasets, our SHD is 135 on the 24V-439N-Microwave dataset, lower than the 147 of DeepSeek, but higher than the 133 of Gemini and the 133 of ChatGPT. It is noteworthy that on the 25V-474N-Microwave dataset, our SHD is 158, higher than the 147 of DeepSeek but lower than the 183 of ChatGPT. LLMs exhibit low SHD because they predict very few edges. For example, DeepSeek achieved a recall of only 0.0584 on the 24V-439N-Microwave dataset, identifying almost no true causal edges, which artificially lowers its SHD.

\begin{tcolorbox}[colback=gray!10, colframe=black, arc=2pt, boxrule=0.5pt, left=6pt, right=6pt, top=6pt, bottom=6pt]
Overall, the proposed method exhibits excellent comprehensive performance in causal discovery from stationary event sequences. On synthetic data, its F1 and SHD substantially outperform existing methods, demonstrating the model efficiency and accuracy under ideal conditions. On real telecommunication network data, RCCD still achieves the highest F1 Score, showing good generalization ability. Although the SHD increases in real-world scenarios, the high precision and high recall together convince us that RCCD can effectively identify causal relationships, providing reliable support for root cause analysis of network alarms.
\end{tcolorbox}

\subsection{Performance Comparison on Observing Incomplete Sequences}
\label{sec:missing}

\begin{figure*}[!htb]

	\centering
	\scriptsize
	\subfloat[15V-30N-Synthetic]{
		\includegraphics[width=0.24\linewidth]{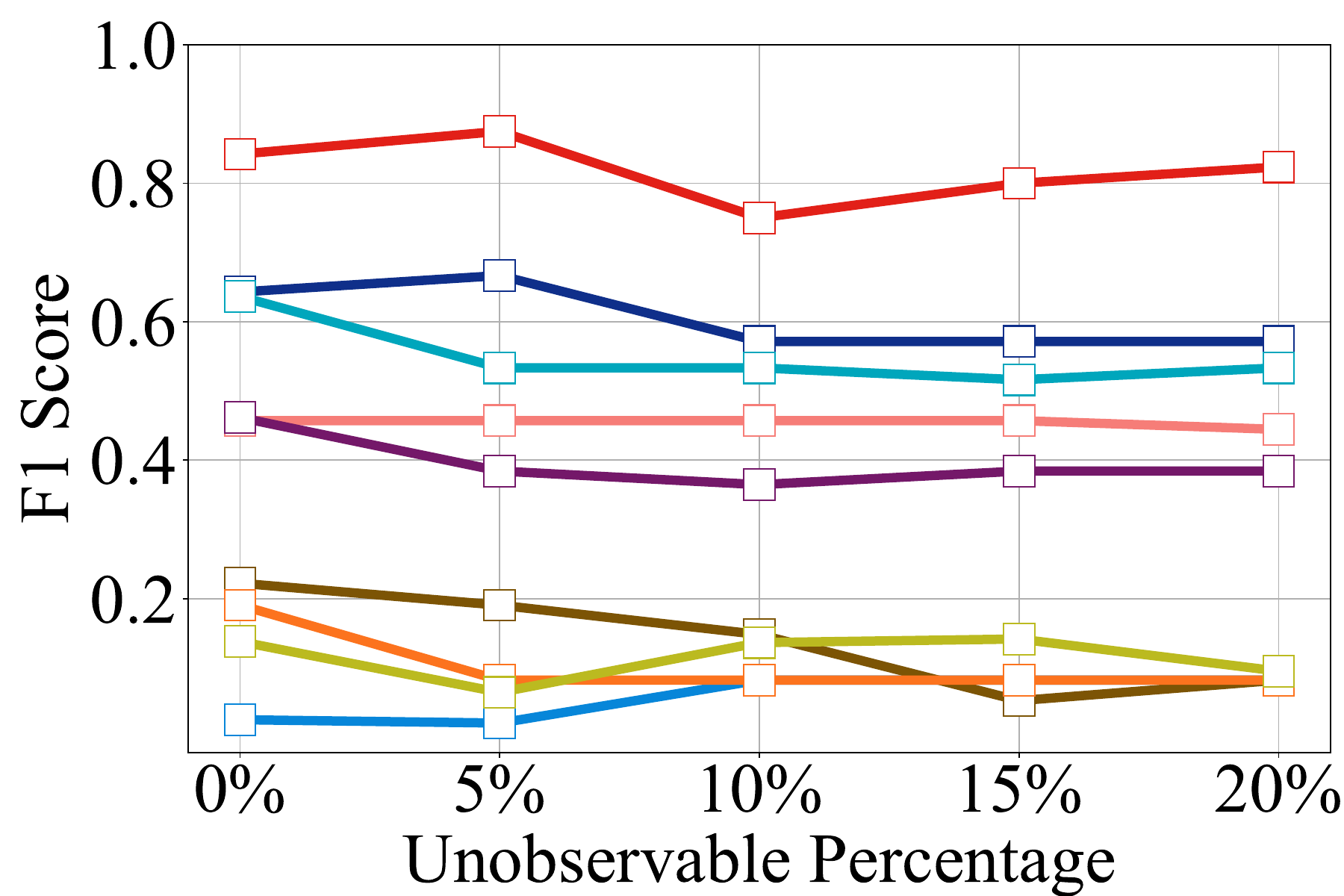}
	}
	\hfill
	\scriptsize
	\subfloat[20V-40N-Synthetic]{
		\includegraphics[width=0.24\linewidth]{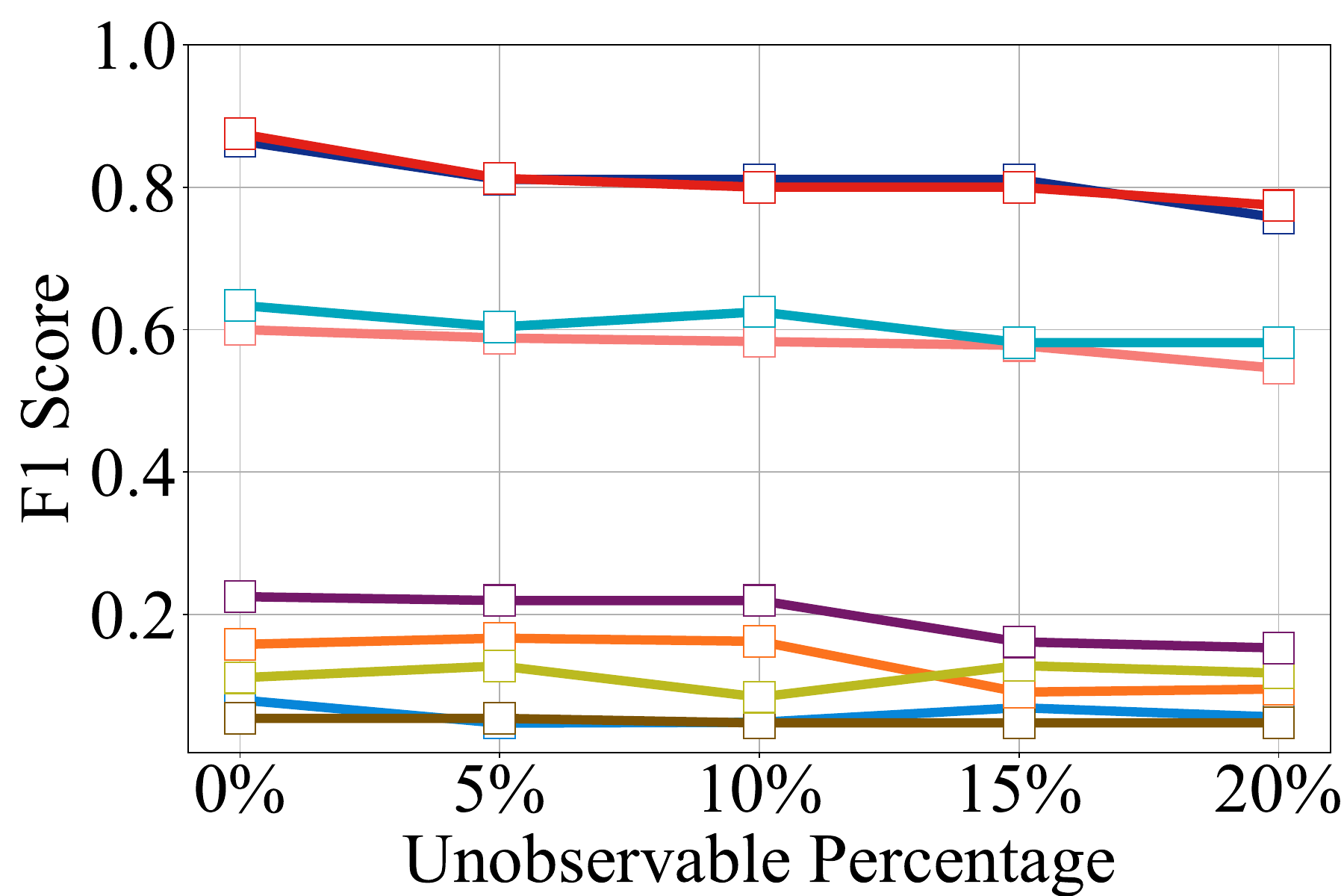}
	}  
	\centering
	\subfloat[24V-439N-Microwave]{
		\includegraphics[width=0.24\linewidth]{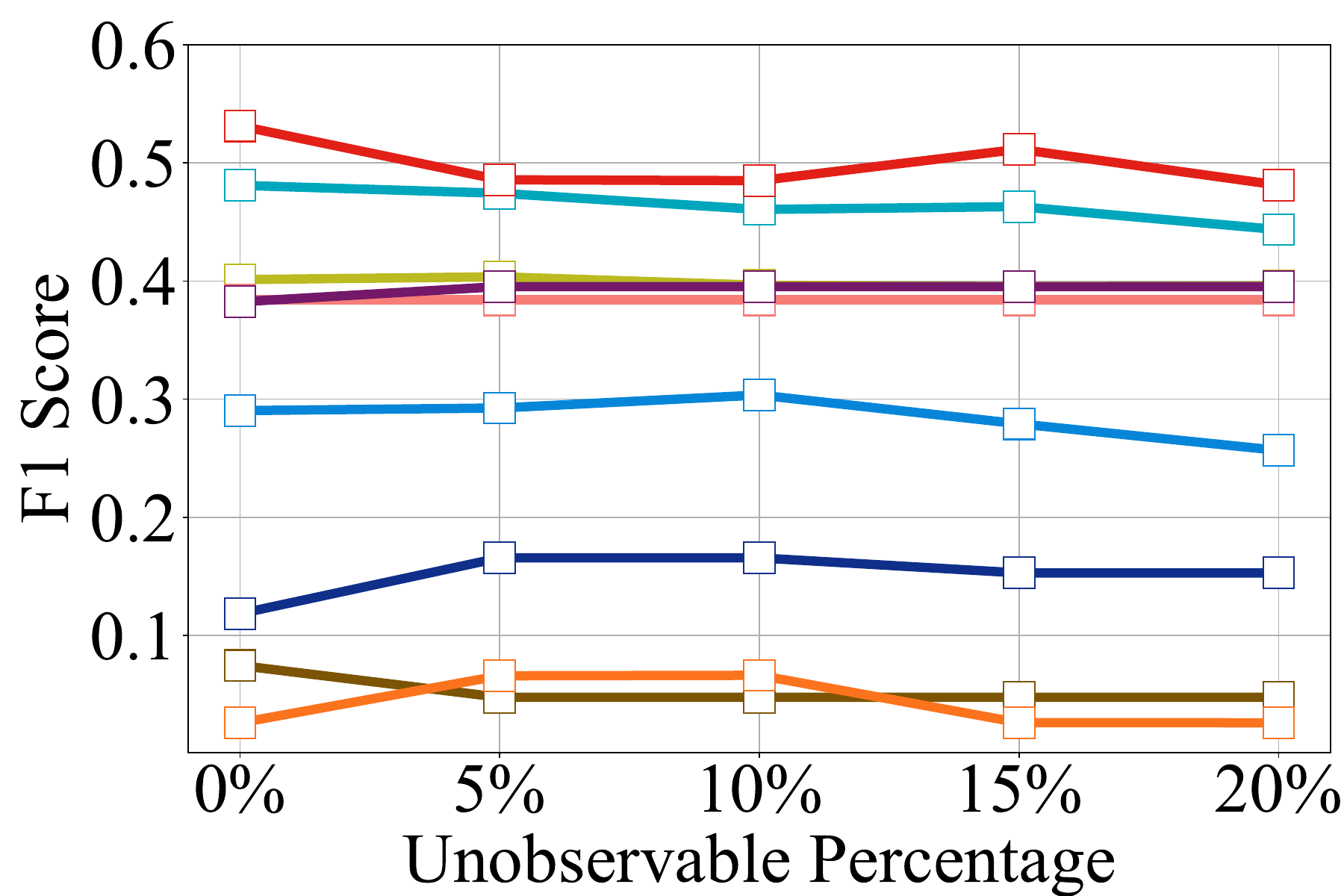}
	}
	\hfill
	\subfloat[25V-474N-Microwave]{
		\includegraphics[width=0.24\linewidth]{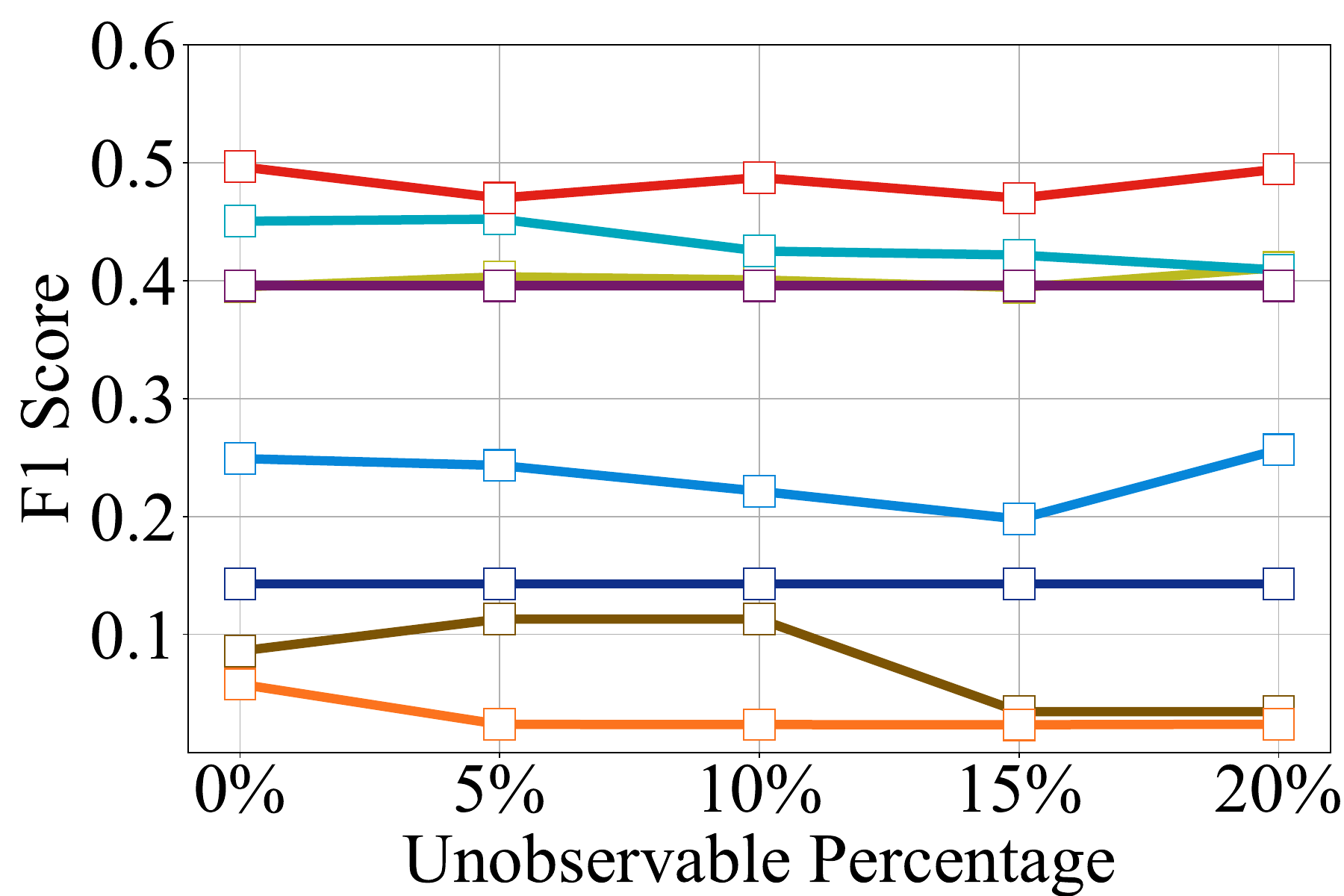}
	}

	\centering
	\scriptsize
	\subfloat[15V-30N-Synthetic]{
		\includegraphics[width=0.24\linewidth]{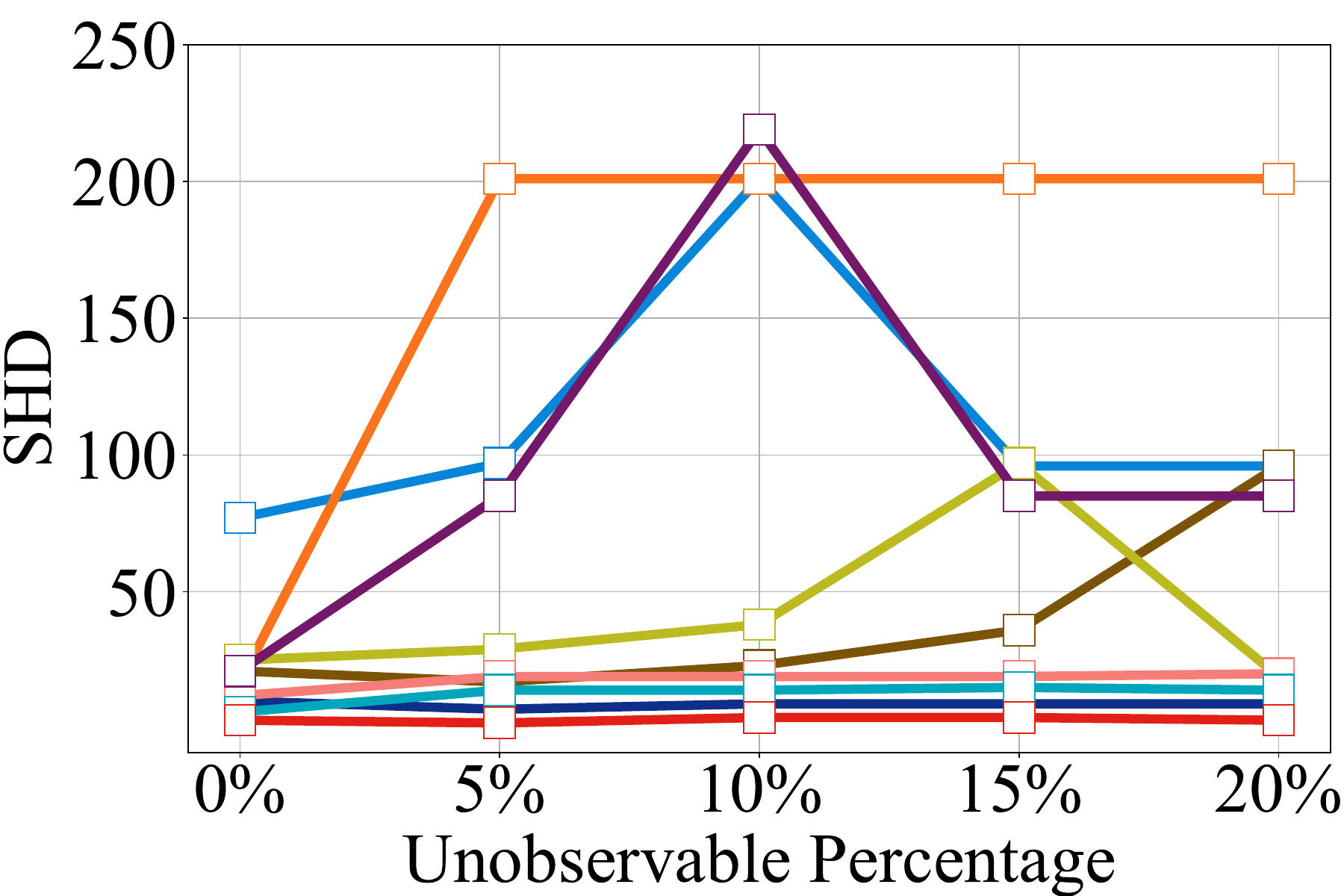}
	}
	\hfill
	\scriptsize
	\subfloat[20V-40N-Synthetic]{
		\includegraphics[width=0.24\linewidth]{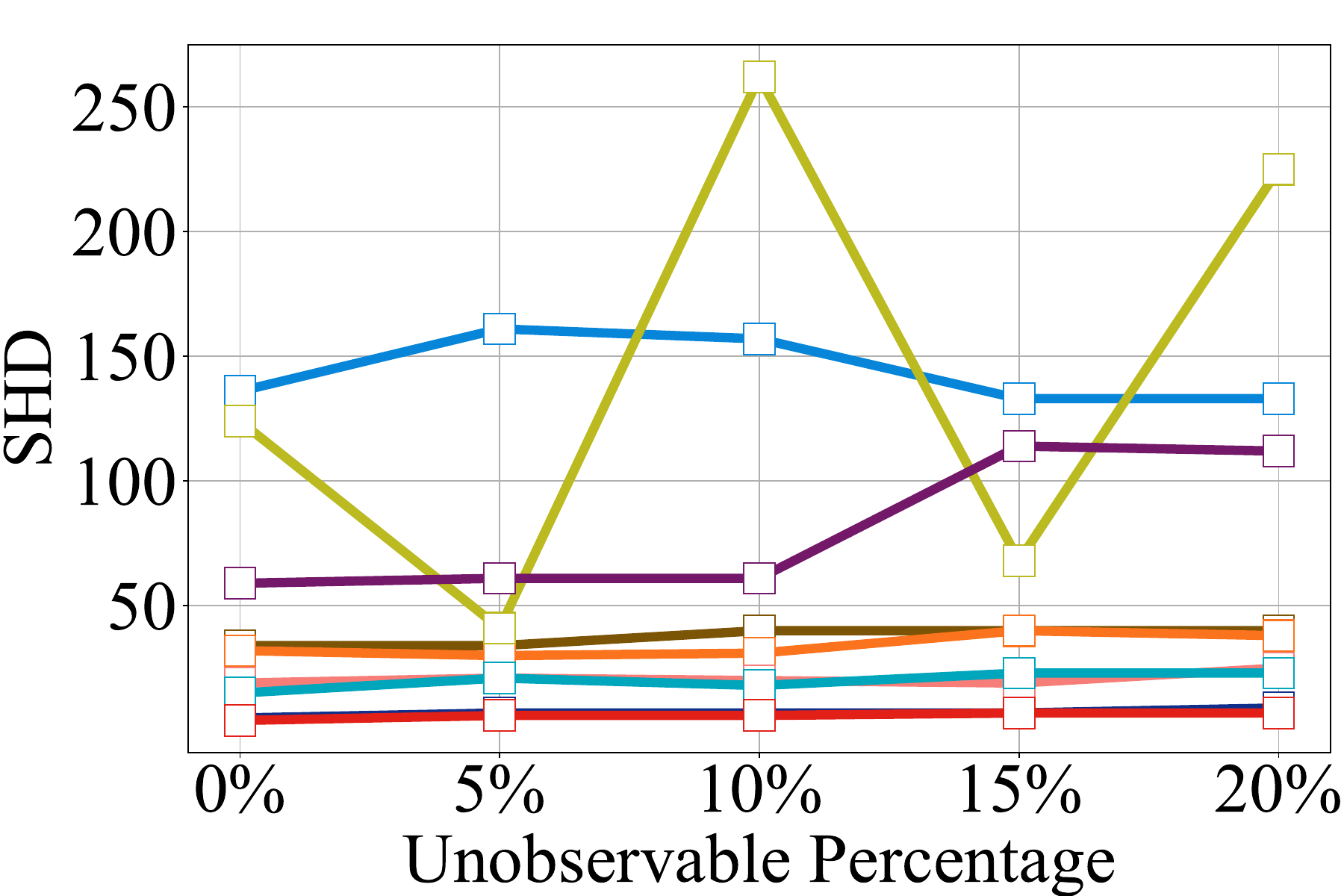}
	}  
	\centering
	\subfloat[24V-439N-Microwave]{
		\includegraphics[width=0.24\linewidth]{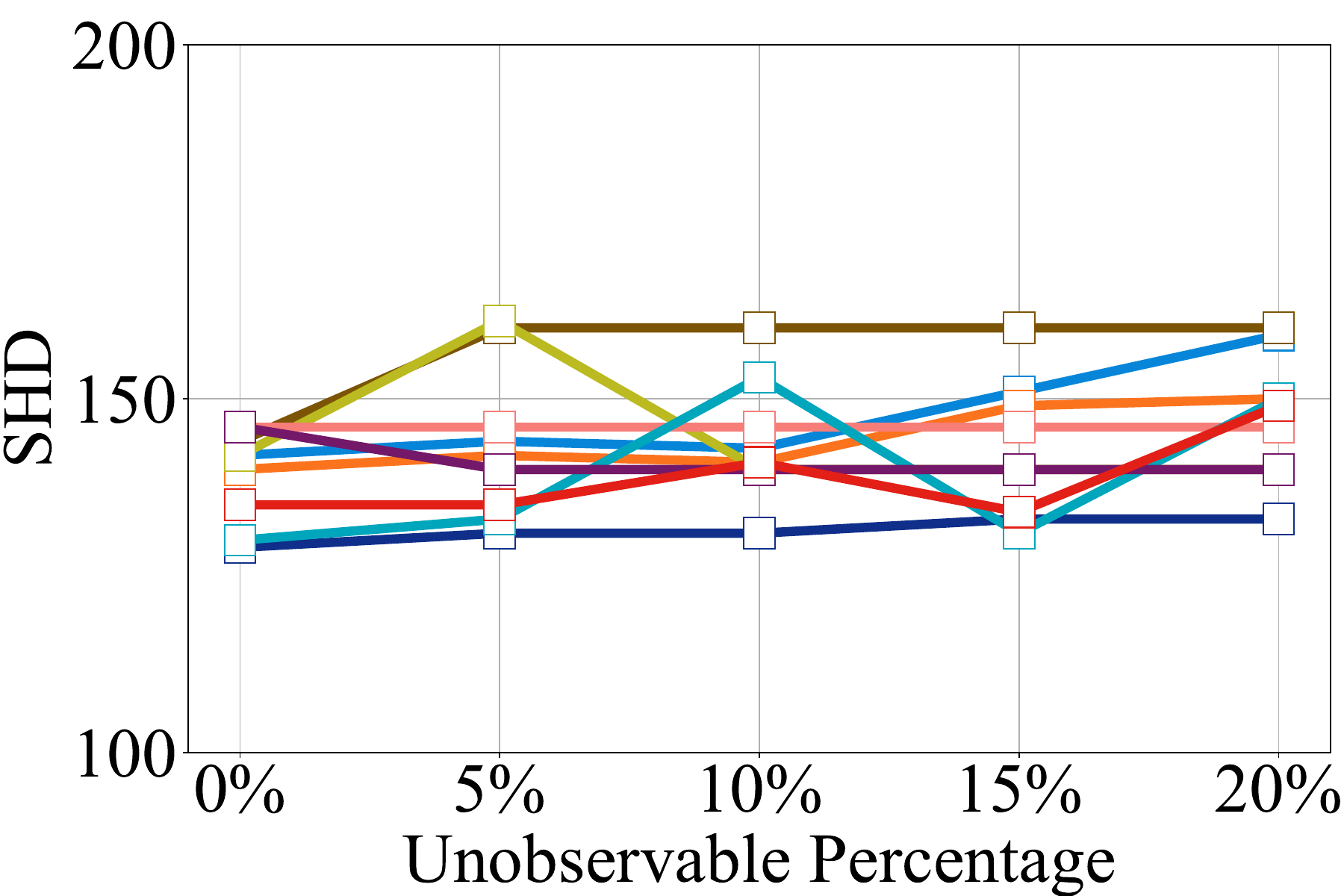}
	}
	\hfill
	\subfloat[25V-474N-Microwave]{
		\includegraphics[width=0.24\linewidth]{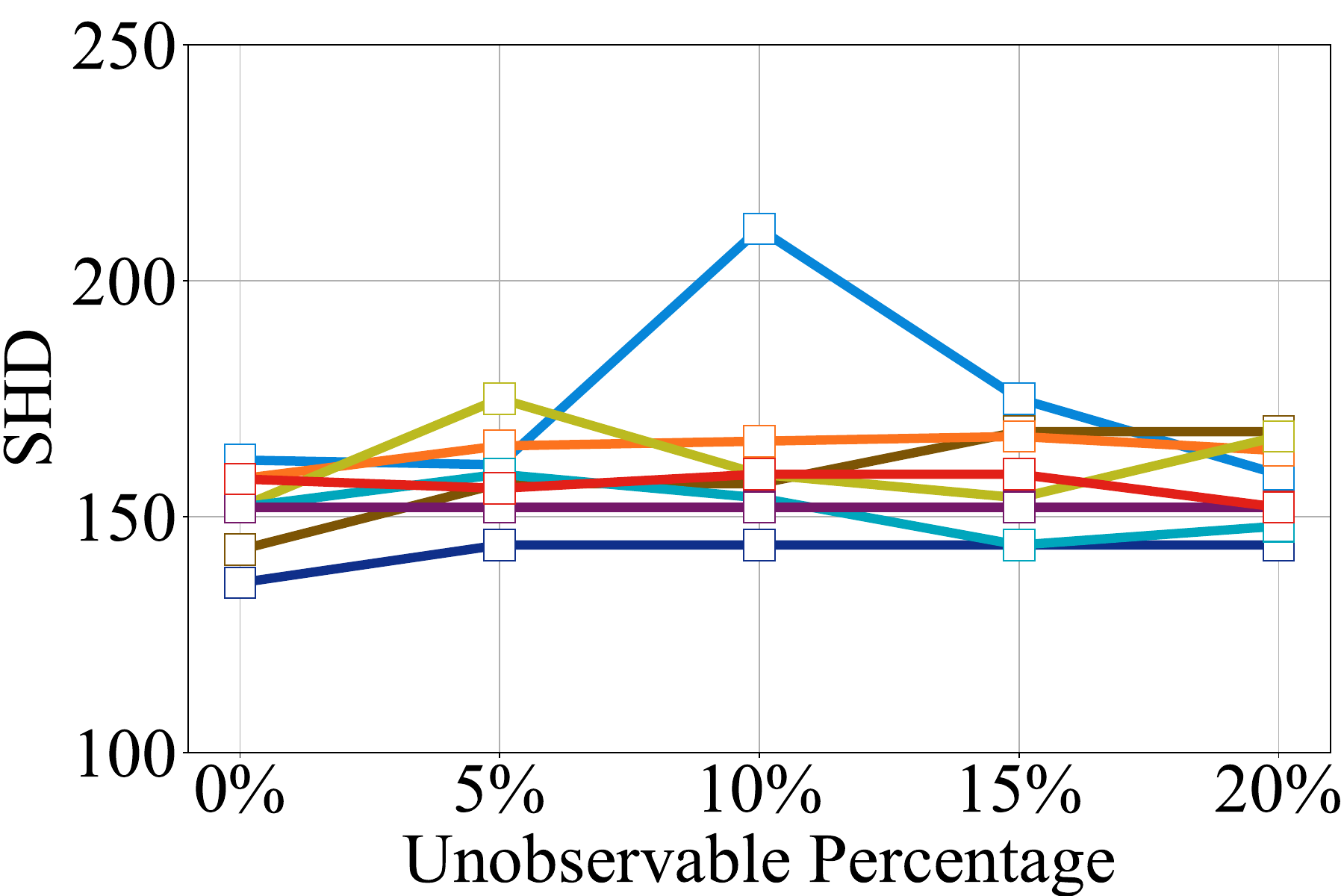}
	}

    \centering
    \scriptsize
	\subfloat{
	\includegraphics[width=0.65\linewidth]{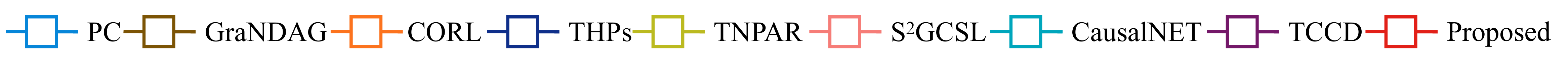}
    }
\caption{Performance Comparison on Observing Incomplete Sequences.}
\label{env:noise}
\end{figure*}

To evaluate the robustness of the proposed method under different noise levels, we simulate random data missing at rates of 5\%, 10\%, 15\%, and 20\% on two synthetic datasets and two real telecommunication network datasets. As shown in Figure \ref{env:noise}, experimental results show that RCCD maintains the highest F1 Score, demonstrating excellent anti-noise capability and robustness.

RCCD leads existing methods under all event missing conditions. On the 15V-30N-synthetic dataset, the F1 is 0.8421 without missing noise. When the missing percentage increases to 0.2, RCCD still achieves an F1 of 0.8235, a very minor decline. In contrast, THPs drops from 0.6429 without missing percentage to 0.5714 at 0.2 missing percentage, and CausalNET drops from 0.6364 to 0.5333. On the 20V-40N-Synthetic dataset, RCCD achieves an F1 of 0.8750 without missing percentage and 0.7742 at 0.2 missing percentage, still much higher than the 0.7568 of THPs and the 0.5818 of CausalNET. On the real dataset 24V-439N-Microwave dataset, RCCD achieves an F1 of 0.5313 without missing percentage and maintains 0.4812 at 0.2 missing percentage, while THPs only reach 0.1529. Although TNPAR has high recall, its extremely low precision leads to an F1 below 0.4. On the 25V-474N-Microwave dataset, RCCD achieves an F1 of 0.4967 without missing percentage and 0.4946 at 0.2 missing percentage, showing almost no effect from data missing. Notably, TNPAR and S$^2$GCSL have recall close to 1 on real datasets but extremely low precision, resulting in F1 Scores far lower than ours. RCCD maintains a good balance between precision and recall, reflecting its strong resistance to missing data. 

Regarding Structural Hamming Distance, RCCD also performs robustly. On the 15V-30N-Synthetic dataset, the SHD is 3 without missing noise and remains 3 at 0.2 missing percentage, far superior to the 9 of THPs, the 14 of CausalNET, and the 85 of TCCD. On the 20V-40N-Synthetic dataset, our SHD increases from 4 without missing percentage to 7 at 0.2 missing percentage, whereas THPs increases from 5 to 9 and CausalNET from 15 to 23. RCCD still achieves the lowest SHD. On the real datasets, the SHD of RCCD increases slightly with missing percentage. However, RCCD maintains a high recall while keeping the SHD within an acceptable range, indicating that the edges it predicts are accurate and cover the true causal structure.

\begin{tcolorbox}[colback=gray!10, colframe=black, arc=2pt, boxrule=0.5pt, left=6pt, right=6pt, top=6pt, bottom=6pt]
Overall, RCCD maintains a leading F1 Score and a stable SHD under different missing percentage levels, exhibiting excellent robustness. Whether on synthetic data or real telecommunication network alarm data, RCCD effectively resists the interference caused by missing data and maintains the best balance between precision and recall. In contrast, existing methods either suffer from low F1 due to over-prediction or have artificially low SHD due to sparse prediction.
\end{tcolorbox}

\subsection{Hyperparameters Sensitivity Analysis}
\label{sec:hparams}

\begin{figure*}[!htb]
	\centering
	\scriptsize
	\subfloat[$\sigma_1$ on 15V-30N-Synthetic]{
		\includegraphics[width=0.24\linewidth]{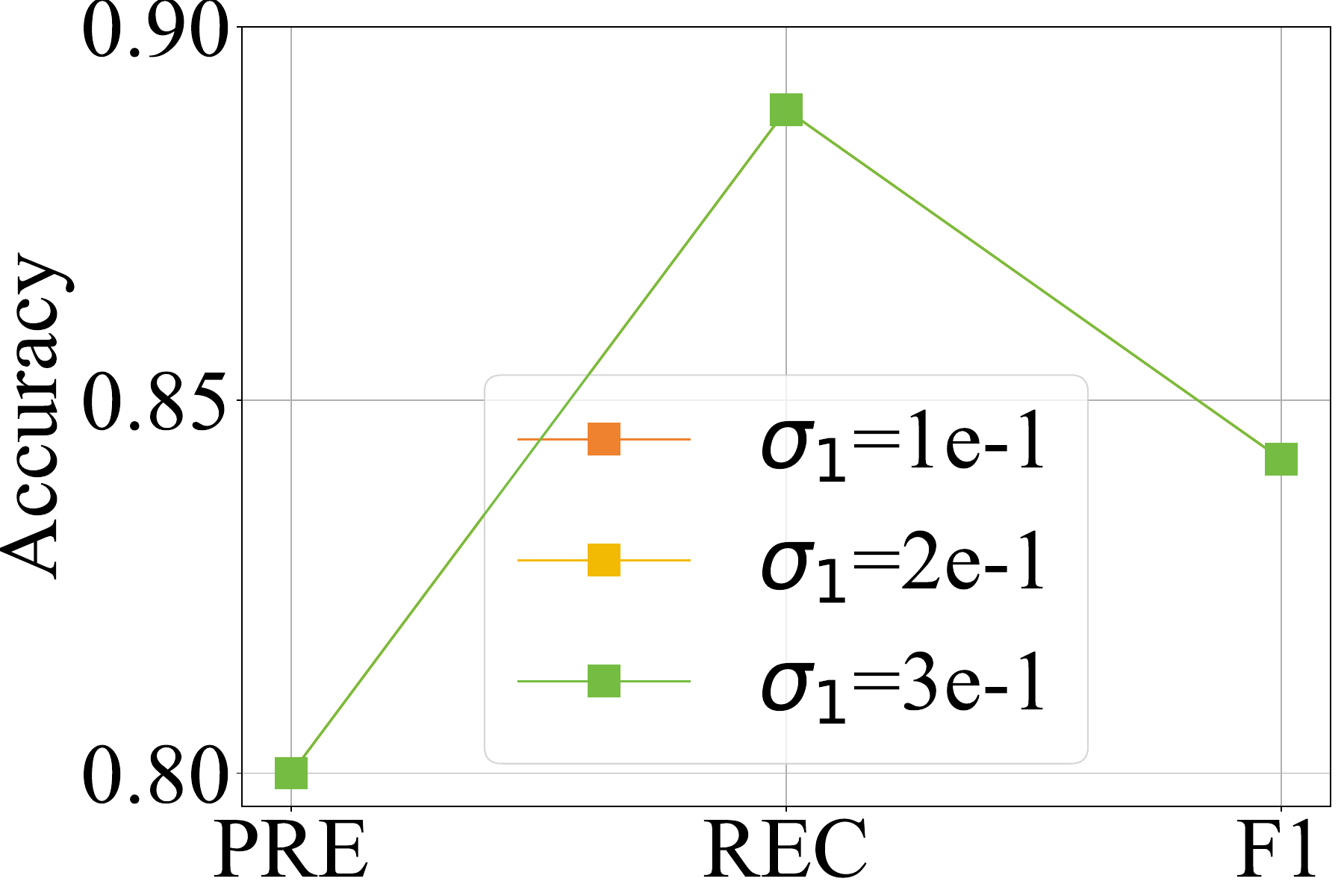}
	}
	\hfill
	\subfloat[$\sigma_1$ on 20V-40N-Synthetic]{
		\includegraphics[width=0.24\linewidth]{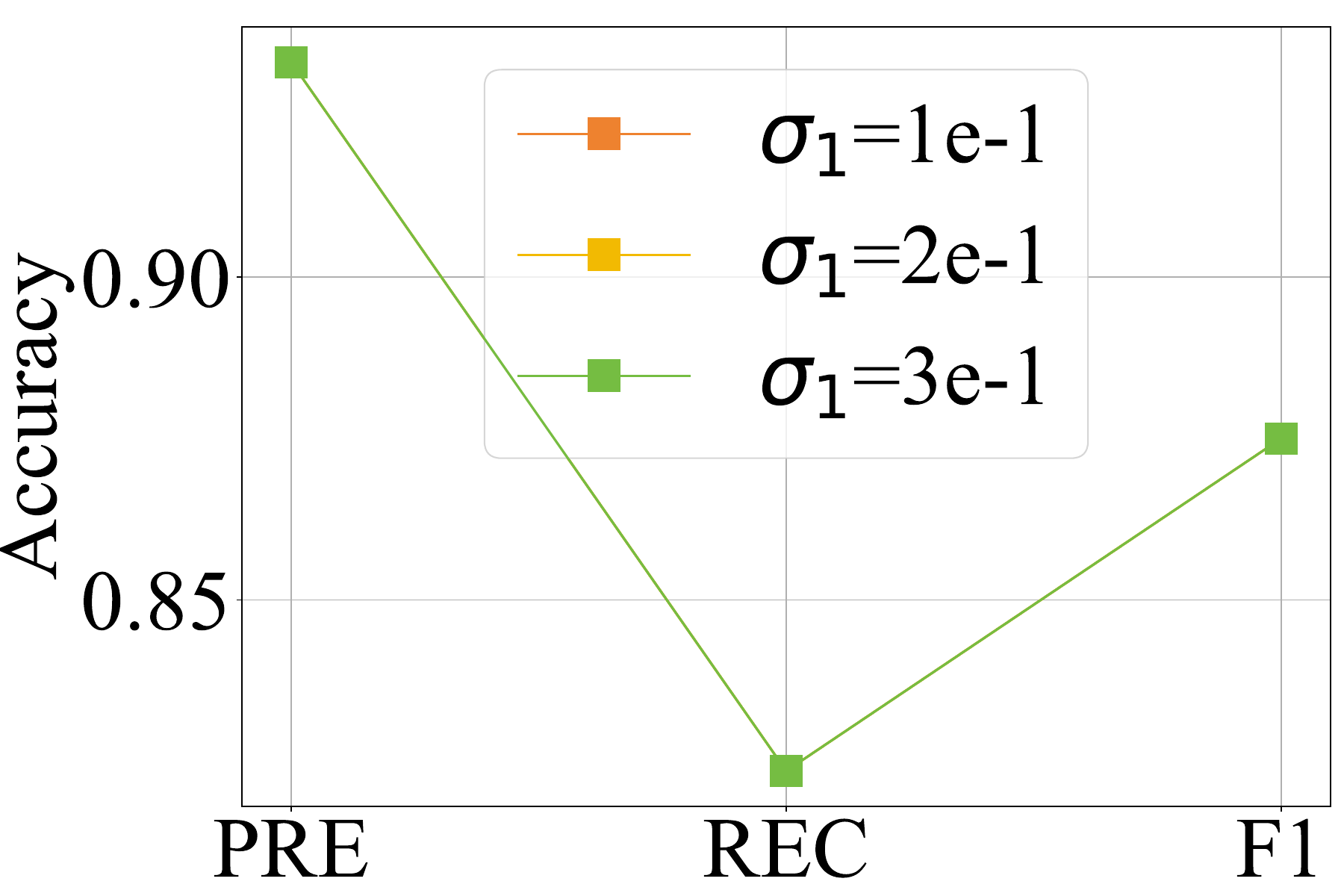}
	}
	\hfill
	\subfloat[$\sigma_1$ on 24V-439N-Microwave]{
		\includegraphics[width=0.24\linewidth]{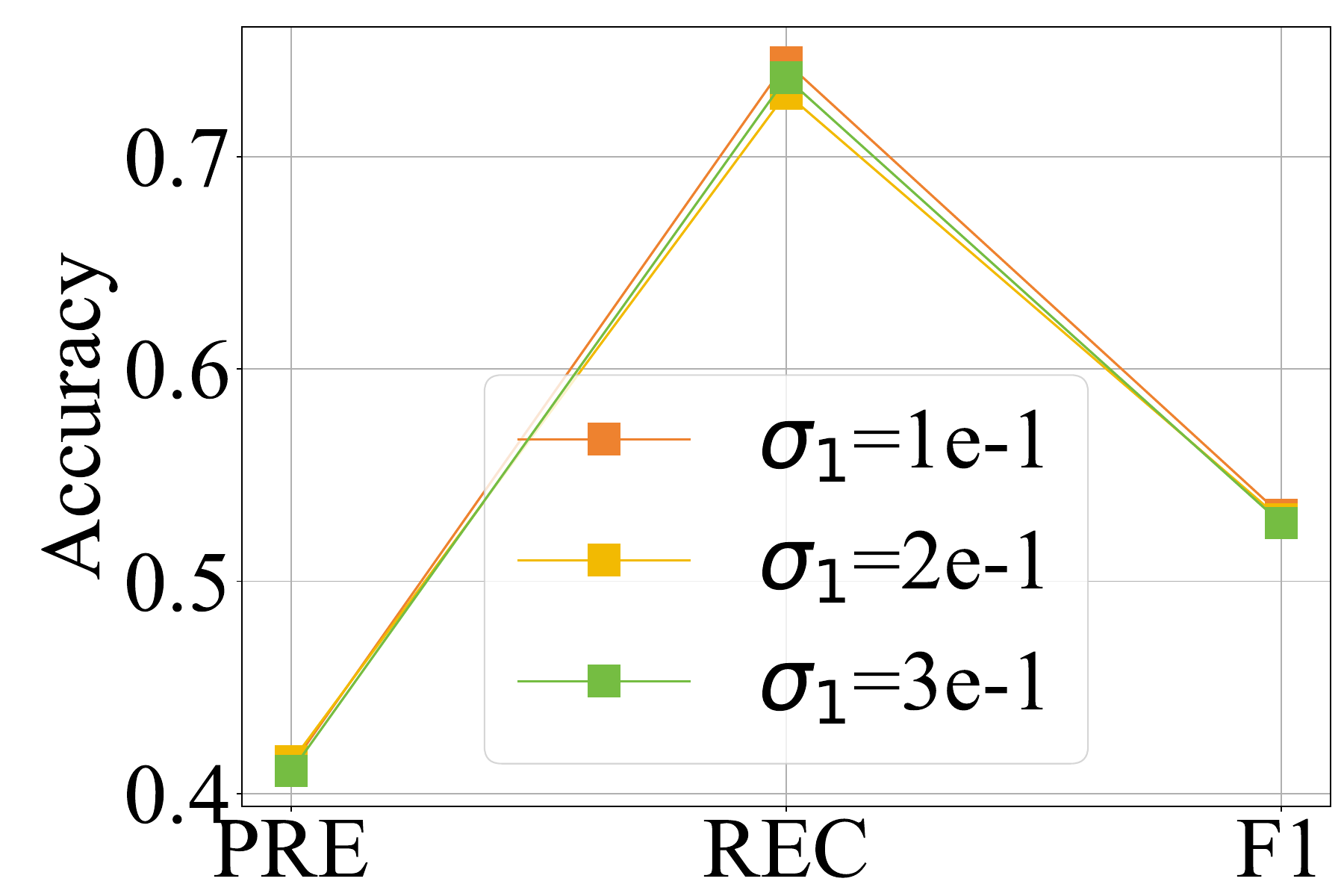}
	}
    \hfill
	\subfloat[$\sigma_1$ on 25V-474N-Microwave]{
		\includegraphics[width=0.24\linewidth]{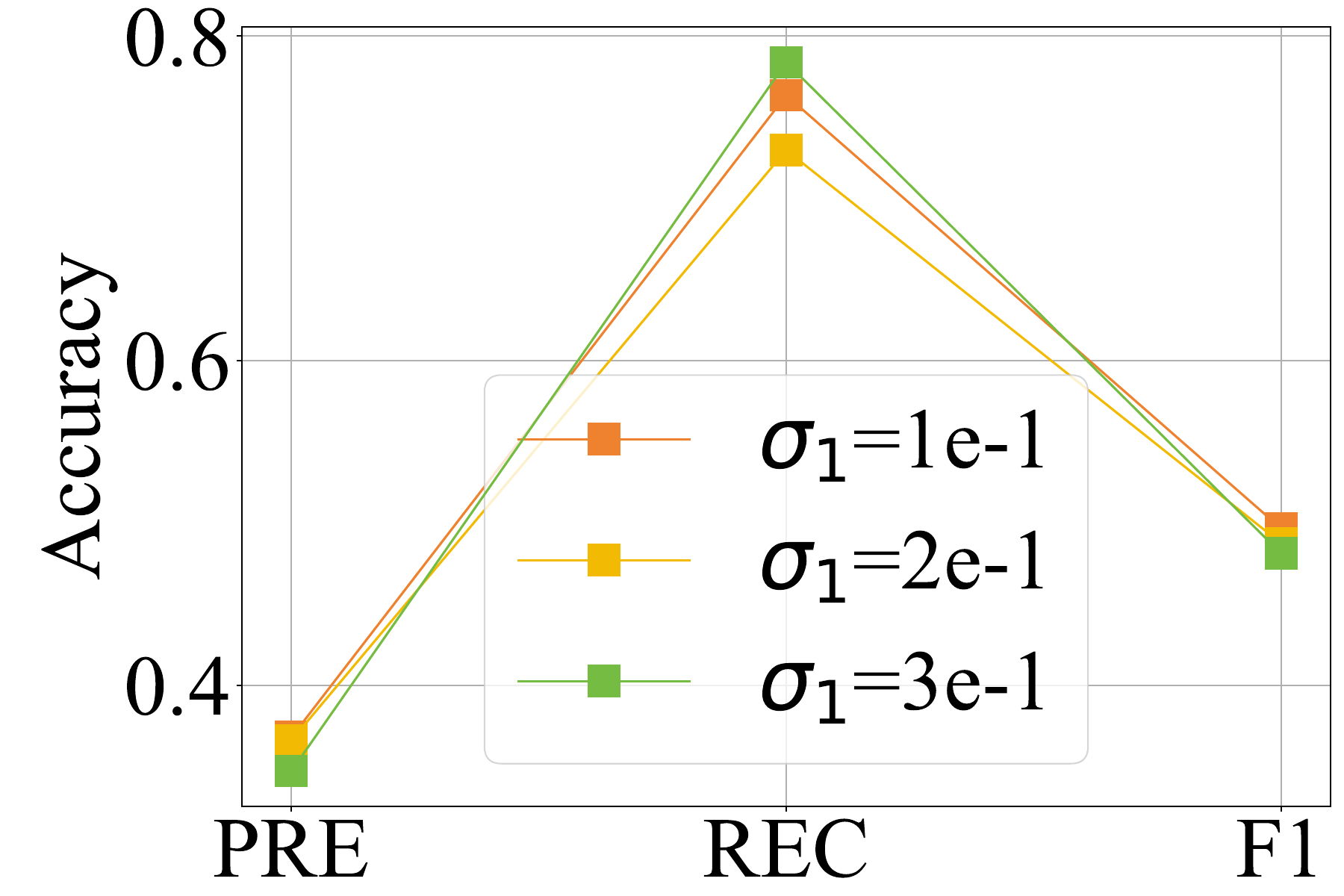}
	}

	\centering
	\scriptsize
	\subfloat[$\sigma_2$ on 15V-30N-Synthetic]{
		\includegraphics[width=0.24\linewidth]{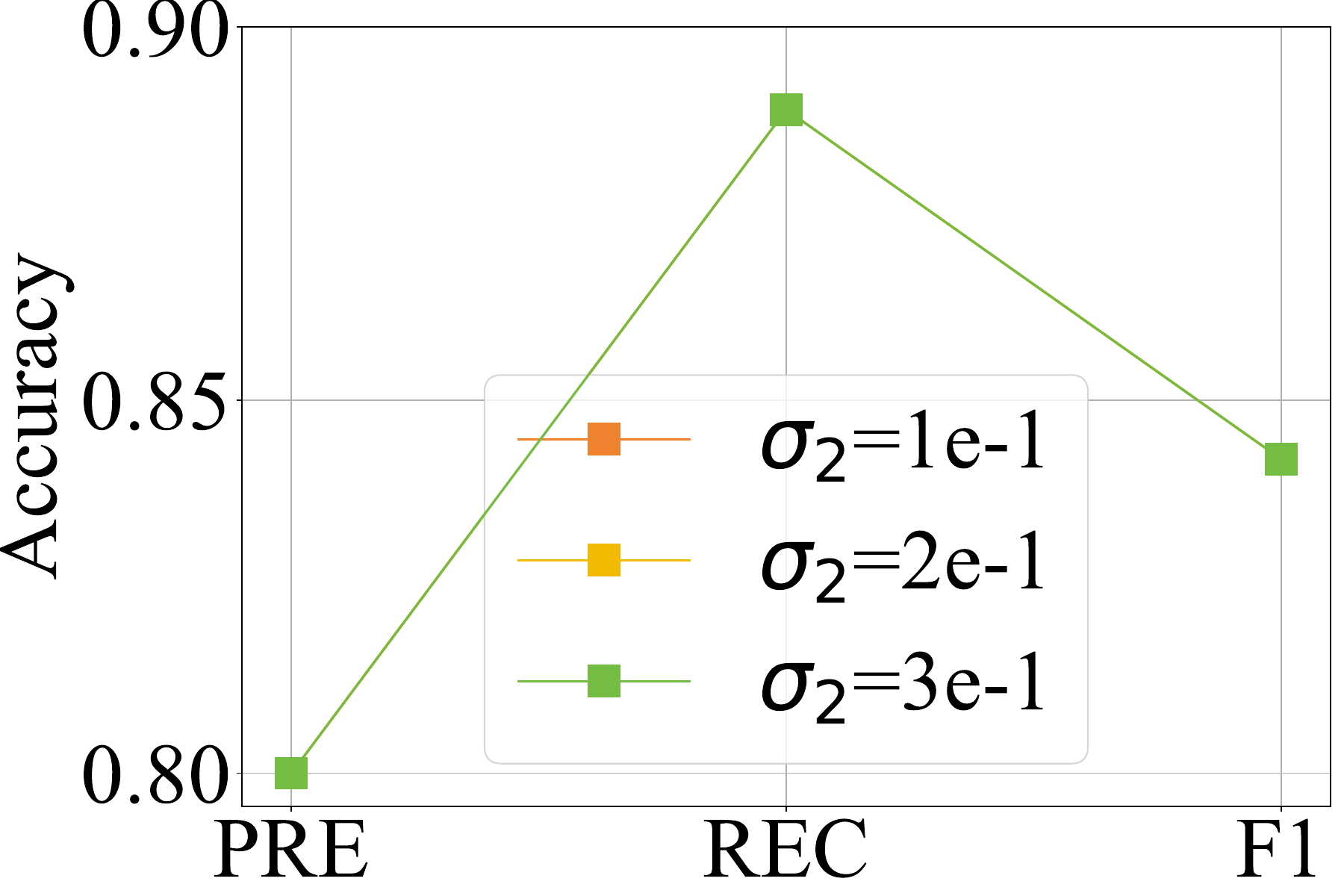}
	}
	\hfill
	\subfloat[$\sigma_2$ on 20V-40N-Synthetic]{
		\includegraphics[width=0.24\linewidth]{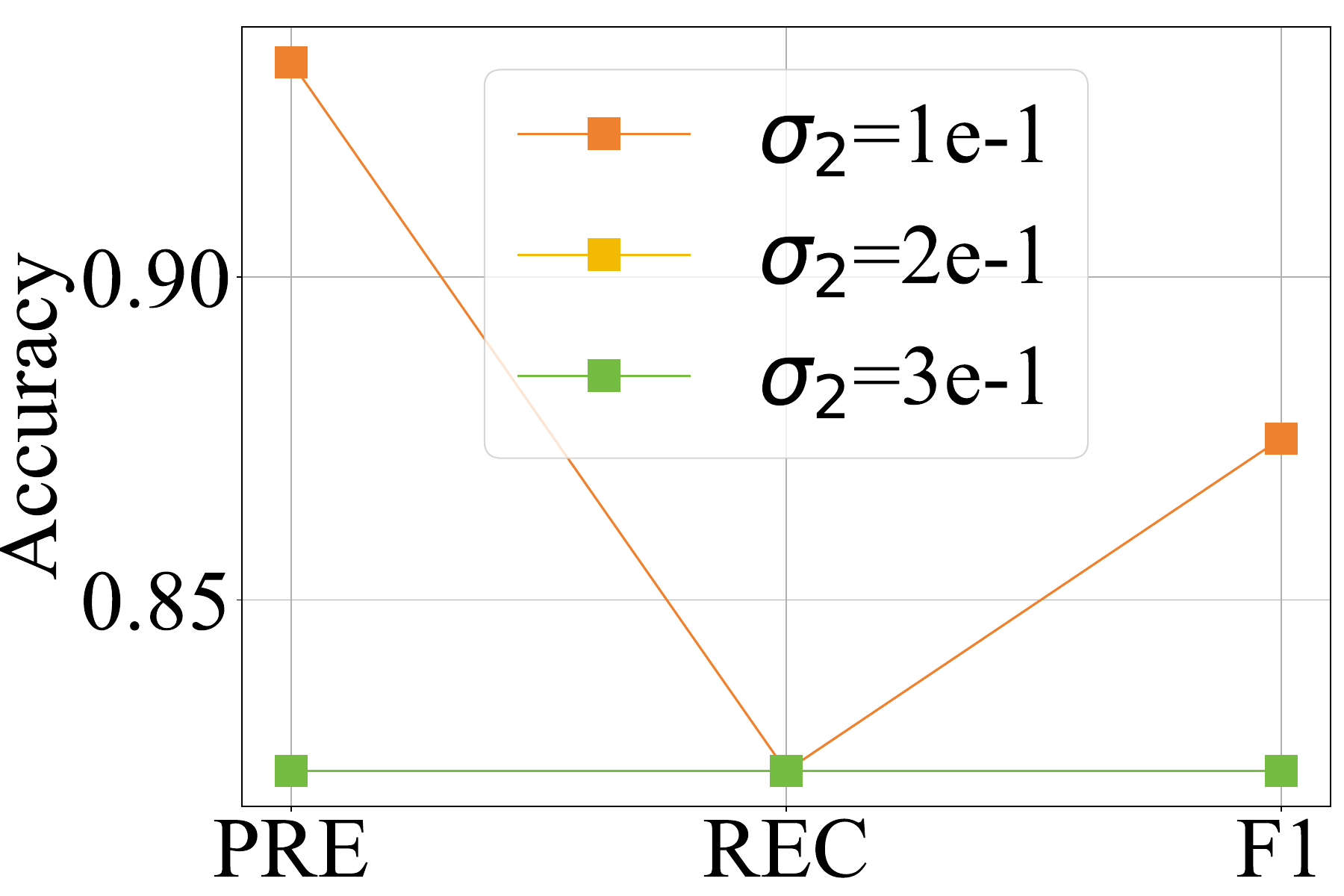}
	}
	\hfill
	\subfloat[$\sigma_2$ on 24V-439N-Microwave]{
		\includegraphics[width=0.24\linewidth]{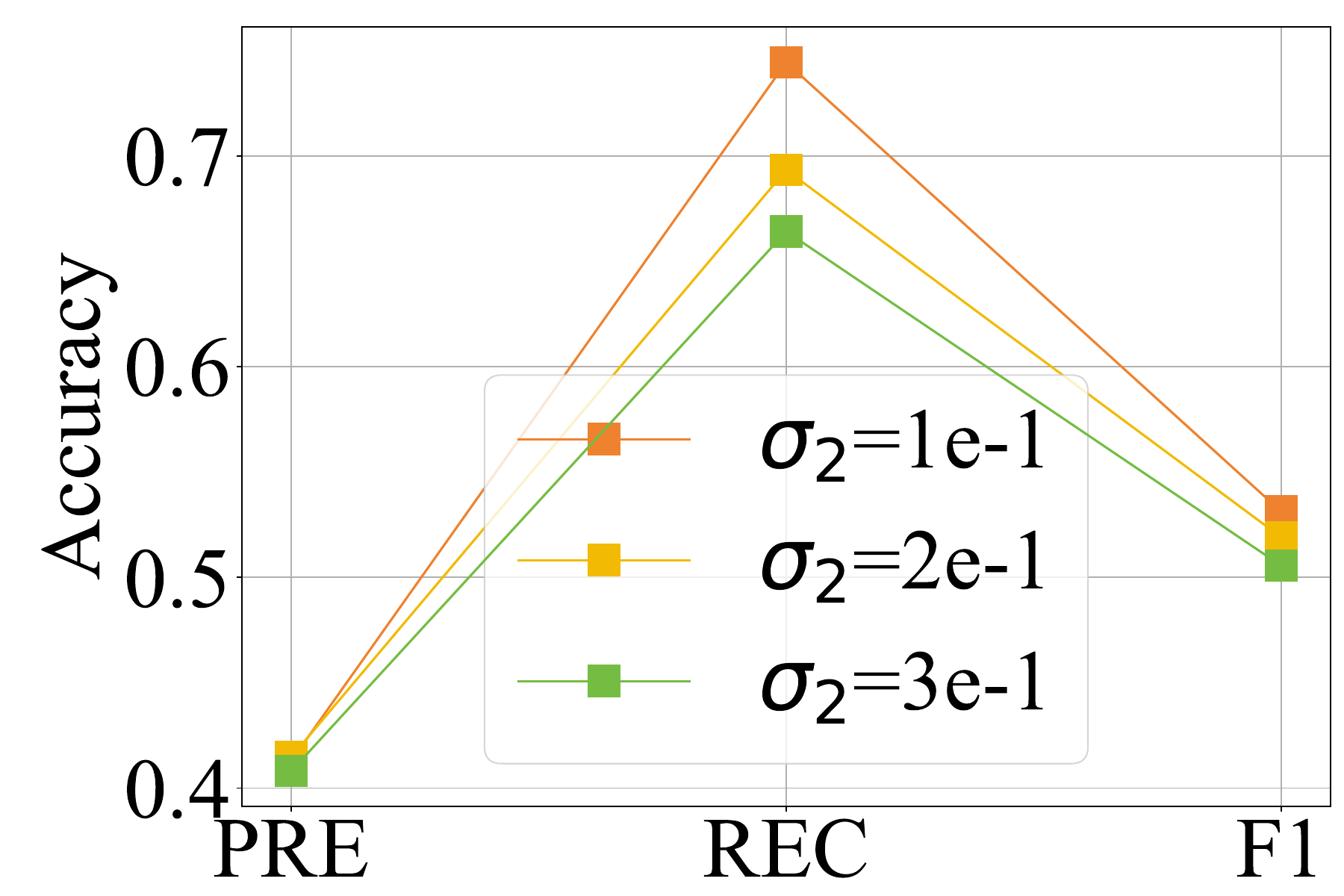}
	}
    \hfill
	\subfloat[$\sigma_2$ on 25V-474N-Microwave]{
		\includegraphics[width=0.24\linewidth]{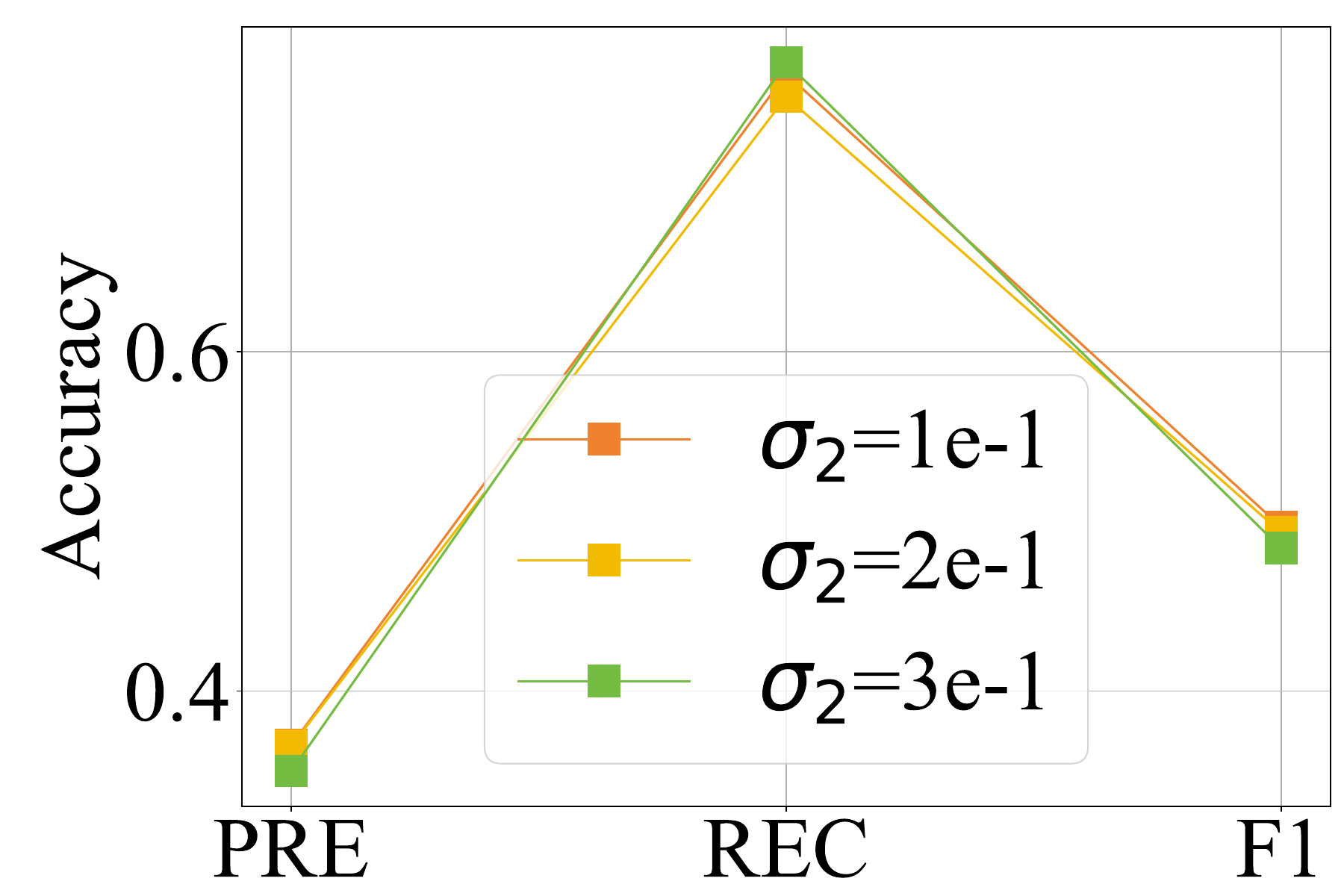}
	}

\caption{Sensitivity to Hyperparameters}
\label{env:sens}
\end{figure*}

To validate the sensitivity of the proposed method to the sparsity regularizer coefficient $\sigma_1$ and the acyclicity regularizer coefficient $\sigma_2$, we set $\sigma_1$ and $\sigma_2$ to 0.1, 0.2, and 0.3 on both synthetic and real-world alarm datasets. As illustrated in Figure \ref{env:sens}, RCCD maintains stable high performance under different parameter values, exhibiting low hyperparameter sensitivity.

For the sparsity regularizer coefficient $\sigma_1$,  the performance of RCCD on synthetic datasets is relatively unaffected by changes in $\sigma_1$. On the 15V-30N-Synthetic and 20V-40N-Synthetic datasets, when $\sigma_1$ increases from 0.1 to 0.3, the precision, recall, F1 Score, and SHD remain unchanged. On the real datasets, changes in $\sigma_1$ cause slight fluctuations. On the 24V-439N-Microwave dataset, the F1 is 0.5313 when $\sigma_1=0.1$, 0.5291 when $\sigma_1=0.2$, and 0.5274 when $\sigma_1=0.3$, a decrease of less than 0.004. On the 25V-474N-Microwave dataset, the F1 is 0.4967 when $\sigma_1=0.1$, 0.4876 when $\sigma_1=0.2$, and 0.4813 when $\sigma_1=0.3$, and the SHD increases from 152 to 160. Experimental results indicate that the sparsity regularizer is insensitive to the value of $\sigma_1$ on synthetic data, and on real data, the F1 decreases slightly as $\sigma_1$ increases but remains stable overall.

For the acyclicity regularizer coefficient $\sigma_2$, RCCD also shows excellent stability. On the 15V-30N-Synthetic dataset, as $\sigma_2$ increases from 0.1 to 0.3, the F1 remains 0.8421, and the SHD remains 3, with no performance degradation. For example, on the 20V-40N-Synthetic dataset, the F1 is 0.8750, and the SHD is 4 when $\sigma_2=0.1$. When $\sigma_2=0.2$, the F1 drops to 0.8235, and the SHD rises to 6, indicating that increasing $\sigma_2$ has some impact on recall on synthetic data, but the decrease in F1 is controllable. On the 24V-439N-Microwave dataset, the F1 is 0.5313 when $\sigma_2=0.1$, 0.5191 when $\sigma_2=0.2$, and 0.5056 when $\sigma_2=0.3$, and the SHD decreases from 135 to 131 and then rises to 135, showing smooth changes. On the 25V-474N-Microwave dataset, the F1 is 0.4967 when $\sigma_2=0.1$, 0.4933 when $\sigma_2=0.2$, and 0.4841 when $\sigma_2=0.3$, and the SHD decreases from 152 to 154 and then rises to 159. Increasing $\sigma_2$ has a small impact on the F1 on real data, and RCCD maintains reasonable causal structure learning ability under different acyclicity constraint strengths.

\begin{tcolorbox}[colback=gray!10, colframe=black, arc=2pt, boxrule=0.5pt, left=6pt, right=6pt, top=6pt, bottom=6pt]
Overall, RCCD exhibits low sensitivity to both the sparsity regularizer coefficient $\sigma_1$ and the acyclicity regularizer coefficient $\sigma_2$. On synthetic datasets, the performance is almost unaffected by parameter changes, and on real datasets, only minor fluctuations occur. Experimental results validate the usability of RCCD, demonstrating that it can achieve stable and efficient performance in causal discovery from event sequences without complex hyperparameter tuning.
\end{tcolorbox}

\subsection{Ablation Study}

\begin{table*}[ht]
\newcommand{\tabincell}[2]{\begin{tabular}{@{}#1@{}}#2\end{tabular}}
\centering
\begin{threeparttable}
\setlength\tabcolsep{2pt}
\caption{Ablation Study}
\begin{tabular}{cccccccccccccccccc}
\toprule  
\multirow{2}*{Scenarios}&\multirow{2}*{Componets}& \multicolumn{4}{c}{15V-30N-Synthetic}& \multicolumn{4}{c}{20V-40N-Synthetic} & \multicolumn{4}{c}{24V-439N-Microwave}& \multicolumn{4}{c}{25V-474N-Microwave}\\
\cline{3-18}~&~&PRE & REC & F1 & SHD& PRE & REC & F1 & SHD& PRE & REC & F1 & SHD& PRE & REC & F1 & SHD\\
\midrule
\multirow{2}*{Stable}&	Baseline&	0.5385&	0.7778&	0.6364&	6&	0.5417&	0.7647&	0.6341&	15&	0.4245&	0.5547&	0.4810&	130&	0.4054&	0.5067&	0.4504&	152\\
~&Baseline+IHCA&	0.8000&	0.8889&	0.8421&	3&	0.9333&	0.8235&	0.8750&	4&	0.4130&	0.7445&	0.5313&	135	&0.3681&	0.7635&	0.4967&	158\\
\bottomrule
\multirow{3}*{Missing}&	Baseline&	0.3810&	0.8889&	0.5333&	14&	0.4211&	0.9412&	0.5818&	23&	0.3185&	0.7299&	0.4435&	150&	0.3937&	0.4256&	0.4090&	148\\
~&Baseline+IHCA&	0.7778&	0.7778&	0.7778&	4&	0.8571&	0.7059&	0.7742&7&	0.2807&	0.8686&	0.4242&	156&	0.3322&	0.6757&	0.4454&	168\\
~&\textbf{IHCA+MACO (RCCD)}&	\textbf{0.8750}&	\textbf{0.7778}&	\textbf{0.8235}&	\textbf{3}&	\textbf{0.8571}&	\textbf{0.7059}&	\textbf{0.7742}&	\textbf{7}&	\textbf{0.3449}&	\textbf{0.7956}&	\textbf{0.4812}&	\textbf{149}&	\textbf{0.3628}&	\textbf{0.7770}&	\textbf{0.4946}&	\textbf{152}\\
\bottomrule
\end{tabular}
\label{tab:ablation_results}
\end{threeparttable}
\end{table*}

To demonstrate the effectiveness of the two novel components (IHCA and MACO) of the RCCD method, we conducted ablation experiments. Using CausalNet as the baseline, we add IHCA and MACO sequentially under both stable event sequences and 20\% missing event sequence scenarios, evaluating the F1 Score and SHD of the method on four datasets. As presented in Table \ref{tab:ablation_results}, experimental results show that adding IHCA alone significantly improves performance, and adding MACO further enhances performance, especially under high noise, verifying that each component positively contributes to the causal discovery task.

Under the stable scenario, the baseline method achieves an F1 of 0.6364 and an SHD of 6 on the 15V-30N-Synthetic dataset. Adding IHCA raises the F1 to 0.8421 and reduces the SHD to 3. On the 20V-40N-Synthetic dataset, the baseline F1 is 0.6341 and SHD is 15. After adding IHCA, the F1 of RCCD rises to 0.8750 and the SHD drops to 4. On the real dataset, 24V-439N-Microwave dataset, the baseline F1 is 0.4810 and SHD is 130. After adding IHCA, the F1 of RCCD rises to 0.5313 and the SHD increases slightly to 135. On the 25V-474N-Microwave dataset, the baseline F1 is 0.4504 and SHD is 152. After adding IHCA, the F1 of RCCD rises to 0.4967 and the SHD slightly increases to 158. Experimental results demonstrate that IHCA brings significant F1 improvements on all datasets.

Under the 20\% event missing scenario, the baseline performance generally declines. On 15V-30N-Synthetic, the baseline F1 drops from 0.6364 to 0.5333 and the SHD rises from 6 to 14. Adding IHCA recovers the F1 to 0.7778 with an SHD of only 4, close to the missing-free level. Adding MACO further achieves an F1 of 0.8235 and an SHD of 3 for RCCD. On 20V-40N-Synthetic, the baseline F1 drops from 0.6341 to 0.5818 and the SHD rises to 23. Adding MACO keeps the F1 of RCCD at 0.7742 with the SHD unchanged. On the 24V-439N-Microwave dataset under noise, the baseline F1 drops to 0.4435 and the SHD is 150. Adding IHCA alone reduces the F1 of RCCD to 0.4242 but the SHD to 156, indicating that IHCA alone introduces too many false positives under real high noise. Adding MACO recovers the F1 of RCCD to 0.4812 and reduces the SHD to 149, matching the noise-free baseline level. On the 25V-474N-Microwave dataset under noise, the baseline F1 is 0.4090 and the SHD is 148. Adding MACO raises the F1 of RCCD to 0.4946 and reduces the SHD to 152, surpassing the noise-free baseline.

\begin{tcolorbox}[colback=gray!10, colframe=black, arc=2pt, boxrule=0.5pt, left=6pt, right=6pt, top=6pt, bottom=6pt]
Overall, both components, IHCA and MACO, contribute positively to performance. IHCA mainly improves recall and F1, while MACO helps maintain structural accuracy and precision under the incomplete observation scenario. Combining these two components achieves optimal performance under all test conditions, validating the complementary effectiveness of the components.
\end{tcolorbox}

\section{Discussion}
\label{sec:7}

\subsection{Why does the RCCD work?}

The effectiveness of the proposed method stems from the collaboration of its two innovative components. First, the influence-aware causal attention mechanism (IHCA) integrates event duration into the embedding representation through a non-linear gated mapping. Through this module, the model can capture the influence of alarm events based on their transience and persistence. Meanwhile, the IHCA mechanism utilizes causal convolution to aggregate local concurrent events into hyper-edges. Subsequently, it injects attention weights by combining topological decay coefficients with a learnable causal weight. Thus, we can capture the complex causal relationships arising from the concurrent effects of multiple events.
In the experiments in Section \ref{sec:stable}, the F1 Scores of RCCD significantly outperform CausalNET and THPs, with SHD as low as 3 and 4, demonstrating its ability to model complex causal relationships. 
Second, the mask-based alternating optimization framework actively hides partial historical events and forces reconstruction in the expectation step, forcing the predictor to learn representations robust to data missing. In the M-step, we fix the predictor to optimize causal graph parameters, and then combine sparse regularization and acyclic regularization to obtain a sparse and acyclic causal graph. Under the extreme missing noise of 20 percent missing events, the F1 Score of the proposed method on the 24V-439N-Microwave dataset only slightly decreases from 0.5313 to 0.4812. In contrast, without the alternating optimization framework with masked imputation, the F1 Score can only reach 0.4242.

Although the proposed method achieves significant improvements in causal discovery accuracy and robustness, certain limitations remain. Presently, the masking strategy relies on predefined random masking probabilities. However, event missingness in telecommunication networks is often non-random. Although fixed probability masking can effectively improve robustness, it may not fully adapt to the dynamic distribution of actual missing data. In future work, we will simulate real collection noise through a learnable masking distribution to further enhance the robust discovery capability of the model for complex causal relationships.

\subsection{Threats to Validity}

During the reproduction of baseline methods, we implemented and compared all competing methods fairly. Specifically, PC, GraNDAG, CORL, and THPs were uniformly implemented based on the gCastle toolchain. S$^2$GCSL and CausalNET were run using their official open-source code. Since TNPAR and TCCD did not release source code, we carefully reproduced their core details and parameter configurations according to the original paper descriptions. However, inconsistencies in reproduction due to differences in text interpretation or parameter tuning deviations remain difficult to avoid completely, which may affect the fairness of the comparison results to some extent.

\section{Conclusion}
\label{sec:8}

In this paper, we proposed a robust concurrent causal discovery method to address the problems of complex causality and missing observations in event sequences. Specifically, we designed an influence-aware causal attention mechanism that incorporates event duration and concurrent information into attention computation, effectively capturing the complex causal relationships triggered by multiple events jointly. Furthermore, we developed a mask-based alternation optimization framework that enhances robustness to missing events through a self-supervised reconstruction task, ensuring reliable causal structure learning under incomplete observations. Extensive experiments on multiple synthetic and real datasets demonstrate that RCCD significantly outperforms existing SOTA methods in terms of discovery performance and robustness. In future work, we plan to explore an adaptive masking mechanism tailored to real-world data to further improve the adaptability of the proposed model to various noise types across different scenarios.

\bibliography{paperbib.bib}
\bibliographystyle{IEEEtran}

\vfill
\end{document}